\pdfoutput=1

\documentclass[11pt]{article}

\usepackage[]{EMNLP2023}

\usepackage{times}
\usepackage{latexsym}

\usepackage[utf8]{inputenc}

\usepackage{microtype}

\usepackage{inconsolata}

\usepackage[T1]{fontenc}
\usepackage{booktabs}
\usepackage{enumitem}

\newcommand{\strict}{\emph{Strict}\xspace}
\newcommand{\smallstr}{\emph{Strict-Small}\xspace}
\newcommand{\loose}{\emph{Loose}\xspace}
\newcommand{\base}{\texttt{base}\xspace}
\newcommand{\dynabench}{Dynabench\xspace}
\newcommand{\hypernym}{\texttt{hypernym}}
\newcommand{\hyponym}{\texttt{hyponym}}
\newcommand{\other}{\texttt{other}}
\newcommand{\n}{\textbackslash n}
\renewcommand{\-}{\hphantom{-}}
\newcommand{\defn}[1]{\textbf{#1}}

\def\best#1{\underline{#1}}
\def\bestall#1{\textbf{\best{#1}}}

\def\contrast#1{\textsc{#1}}
\def\testsuite#1{\textsc{#1}}

\usepackage[disable]{todonotes}
\usepackage{array}
\usepackage{multirow,graphicx}
\usepackage{float}
\usepackage{cleveref}
\usepackage{xspace}
\usepackage{capt-of}

\title{Findings of the \includegraphics[height=0.4cm]{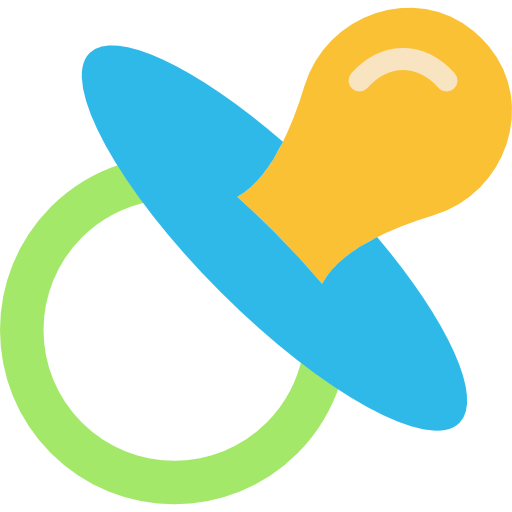} BabyLM Challenge:\\Sample-Efficient Pretraining on Developmentally Plausible Corpora}

\author{
         Alex Warstadt$^{1*}$ \hfill Aaron Mueller$^{2,3*}$ \hfill Leshem Choshen$^{4,5}$ \hfill \textbf{Ethan Wilcox}$^1$ \hfill    \textbf{Chengxu Zhuang}$^4$\\
        \textbf{Juan Ciro}$^6$ \ \ \ \ \ \ \  \textbf{Rafael Mosquera}$^6$\ \ \ \ \ \ \ \textbf{Bhargavi Paranjape}$^8$\\
        \textbf{Adina Williams}$^{6,7}$\ \ \ \ \ \ \ \textbf{Tal Linzen}$^9$\ \ \ \ \ \ \ \textbf{Ryan Cotterell}$^1$\\
        $^1$ETH Zürich\ \ \ \ \ \ \ $^2$Northeastern University\ \ \ \ \ \ \ $^3$Technion \ \ \ \ \ \ \
        $^4$MIT \\ $^5$IBM Research \ \ \ \ \ \ $^6$MLCommons \ \ \ \ \ $^7$Meta AI (FAIR) \\
        $^8$University of Washington \ \ \ \ \ $^9$New York University\\
        \texttt{\href{mailto:warstadt@ethz.ch}{warstadt@inf.ethz.ch}}\quad \texttt{\href{mailto:aa.mueller@northeastern.edu}{aa.mueller@northeastern.edu}}
}
\begin{document}
\maketitle
\def\thefootnote{*}\footnotetext{Equal contribution.}
\renewcommand*{\thefootnote}{\arabic{footnote}}

\begin{abstract}
Children can acquire language from less than 100 million words of input. 
Large language models are far less data-efficient: they typically require 3 or 4 orders of magnitude more data and still do not perform as well as humans on many evaluations. 
These intensive resource demands limit the ability of researchers to train new models and use existing models as developmentally plausible cognitive models.
The BabyLM Challenge is a communal effort
in which participants compete to optimize language model training on a fixed data budget. 
Submissions are compared on various evaluation tasks targeting grammatical ability, downstream task performance, and generalization.
Participants can submit to up to three tracks with progressively looser data restrictions.
From over 30 submissions,
we extract concrete recommendations on how best to train data-efficient language models, and on where future efforts should (and perhaps should not) focus.
The winning submissions 
using the LTG-BERT architecture \citep{samuel2023trained} outperformed models trained on trillions of words.
Other submissions achieved strong results through training on shorter input sequences or training a student model on a pretrained teacher.
Curriculum learning attempts, which accounted for a large number of submissions, 
were largely unsuccessful, though some showed modest improvements.\looseness=-1
\end{abstract}

\section{Introduction}

Although there have massive improvements in the effectiveness of neural language models in the last decade, humans are still the state of the art in language learning. 
To achieve impressive results, language models need to be trained on hundreds of times more language input than a typical human will be exposed to in an entire lifetime. 
The BabyLM Challenge is a shared task that invites members of the natural language processing, linguistics, and cognitive science communities to train language models in low-resource data settings, where the amount of linguistic input resembles the amount received by human language learners. 
In doing so, our motivations (\Cref{sec:goals}) are to improve the relevance of language models as cognitive models of human language acquisition, find more effective and data-efficient training algorithms for language models, and democratize research on language model training by emphasizing research questions that can be addressed on a smaller training budget.\looseness=-1

\begin{figure}
    \centering
    \includegraphics[width=\linewidth]{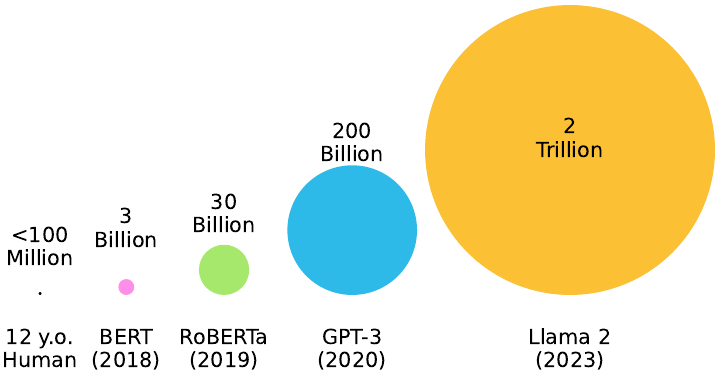}
    \caption{\textbf{Data Scale:} Modern Language Models are trained multiple orders of magnitude more word tokens than the amount available to a typical child. This image is based on Fig.~1 from \citet{warstadt2022artificial}.}
    \label{fig:data-scale}
    \vspace{-10pt}
\end{figure}

Participants in the shared task could submit to the \strict, \smallstr, or \loose track, which, respectively, required models to be trained on corpora that constituted either 10 million words, 100 million words, or 100 million words plus an unlimited amount of additional non-linguistic data (\Cref{sec:instructions}). 
These corpora were constructed from a mixture of sources including developmentally plausible domains such as child-directed speech, transcribed dialogue, and children's literature (\Cref{sec:data}).
To enable standardized evaluation and easy comparison of the resulting models, we create a leaderboard and release an evaluation pipeline (\Cref{sec:evaluation}) targeting zero-shot grammatical performance, finetunability on language understanding tasks, and model inductive bias.
We also contribute a novel set of zero-shot evaluation tasks targeting semantic and discourse-level phenomena.

We received 31 papers making a variety of contributions, ranging from designing novel architectures and tuning hyperparameters to employing curriculum learning and training teacher--student model pairs (\Cref{sec:submissions}). 
We conduct a meta-analysis of the results, yielding several concrete recommendations and scientific conclusions (\Cref{sec:results}).
The winners of the challenge's various tracks made contributions that led to impressive improvements in our evaluation over not just the BabyLM baselines, but also the massively pretrained Llama 2 model \citep{Llama2}.
The best-performing models overall \citep{babylm2023notall} use the LTG-BERT architecture \citep{samuel2023trained}, which synthesizes a number of recent optimizations of the Transformer architecture.
The winner of the \loose track \citep{babylm2023contextualizer} trains the models continuously on the training samples belonging to the same source dataset while randomizing the dataset orders in each training epoch.
Other submissions did not achieve strong downstream results, but still provided valuable scientific contributions. 
We received many curriculum learning submissions,  including one that systematically tested a variety of strategies \citep{babylm2023climb} and reported few improvements over non-curriculum baselines.
\citet{babylm2023gptlike} found that benchmark performance is not correlated with a greater ability to predict human psycholinguistic data.

We plan to organize future BabyLM Challenges that will build on the success of this first iteration (\Cref{sec:future}).
The winning submission from this year sets a high baseline for next year. 
Future iterations will need harder and more varied evaluations, including those that emphasize human-like processing and learning; they should emphasize new approaches that were not thoroughly explored this year, such as multimodality; and, they should incentivize compute-efficiency.
Altogether, the first BabyLM Challenge has been a successful initiative, and we hope that this will continue to advance research on small-scale language models.

\looseness=-1

\section{Motivation}\label{sec:goals} 

The observation at the center of the BabyLM Challenge is this: Children are incredibly data-efficient language learners, and language models are not. 
Children are exposed to less than 100 million word tokens by age 13 \citep{gilkerson2017mapping}, while modern language models are typically trained on 3 or 4 orders-of-magnitude more data (Figure \ref{fig:data-scale}). 
This discrepancy raises two important questions:
First, how is it that humans are able to learn language so efficiently?
Second, what insights from human language learning can be used to improve language models?

A great deal of recent work in language model training seeks improvements by scaling up pretraining data and parameters \citep{raffel-etal-2020-t5,brown-etal-2020-gpt3,hoffmann2022training,Llama2}.
Scaling is undoubtedly central to building deployable models (though see \citealt{mckenzie2023inverse} for counterexamples) and raises its own set of scientific questions, such as quantitative scaling laws \citep{kaplan2020scaling} and the emergence of new abilities \citep{wei2022emergent}.
However, increased emphasis on scaling is unlikely to lead to answers to the two questions we raised, and it excludes researchers without access to massive computational resources.\looseness=-1

Thus, there are three principal benefits to data-limited language model training which the BabyLM Challenge aims to highlight: 
\begin{enumerate}
    \item Building more cognitively and developmentally plausible models of human language acquisition and processing,
    \item Optimizing training pipelines prior to scaling by allowing for faster iteration on architectures and hyperparameters, and
    \item Enabling research on language model training beyond highly funded industry groups.
\end{enumerate}

\paragraph{Cognitive Modeling.}

Language models have been used to model aspects of human language learning and processing for decades \citep[o.a.]{elman1990finding,hale2001probabilistic,reali2005uncoveringa}. 
While many researchers continue to advocate for language models as cognitive models \citep{keller-2010-cognitively,dupoux-2018-cognitive,linzen2019can,baroni-2022-proper,warstadt2022artificial,piantadosi-2023-modern,wilcox2023using}, most agree that it is critical to make LMs learn in more human-like ways. 
\citet{warstadt2022artificial} and \citet{linzen-2020-accelerate} point to data quantity as the most egregious advantage that modern language models have over humans.
When restricted to developmentally plausible data volumes, language models no longer perform well on benchmarks for human-like syntactic and semantic behavior \citep{vanschijndel2019quantity,zhang2021when}.\looseness-1

Working to close the data-efficiency gap between language models and humans will have two principal advantages for cognitive modeling. First, by reverse-engineering known and hypothetical aspects of the human learning scenario---from multimodal inputs and multi-agent interaction to innate linguistic structural biases---we can determine which factors are critical to our unique ability to learn language efficiently \citep{dupoux-2018-cognitive}. Second, by minimizing differences between humans and models, we make results from controlled experiments carried out on models more likely to be applicable to humans \citep{warstadt2022artificial}.\looseness=-1

\paragraph{Faster iteration on architectures and hyperparameters for language modeling.}


Reducing the scale of training provides researchers with a sandbox in which to more fully explore this design space and better optimize training pipelines.
The search space for design choices when training language models is enormous. 
Thus, it can be impractical, especially at large scales, to experiment with new model architectures, training objectives, or data preprocessing steps, in addition to necessary hyperparameter tuning. 
Models such as RoBERTa \citep{liu-etal-2019-roberta} have succeeded in making some optimizations to the BERT training pipeline, but more optimizations remain. Indeed, there are anecdotes of basic design choices for popular pipelines, such as the masking rate for BERT training \citep{wettig-etal-2023-mask}, being poorly tuned for years, despite hundreds or even thousands of papers using this training pipeline.\looseness=-1

There are numerous dimensions along which to scale down training. Some works seek to optimize pipelines for a limited amount of compute, time, or money. Notable examples of such pipelines for bidirectional encoder-only include ELECTRA \citep{clark2020electra}, 24-hour BERT \citep{izsak2021how}, and MosaicBERT \cite{portes2023mosaicbert}. These pipelines typically combine multiple approaches, such as modifying training objectives to increase the number of supervised predictions per forward pass, using low-precision floating-point computations for certain components, reducing sequence length or padding, and altering the attention or feed-forward layers of the transformer block.

However, the objective of optimizing pipelines for a fixed data budget is relatively underexplored.
This is changing in the last year with new models optimized for small datasets such as LTG-BERT \citep{samuel2023trained} and community-oriented events centered around data-limited training such as the Learning from Small Data workshop \citep{breitholtz2023proceedings} and the MiniPile Challenge \citep{kaddour2023minipile}.\looseness=-1



\paragraph{Democratizing language model training research.}

The third goal of the BabyLM Challenge is to democratize research on pretraining---typically thought to be practical only for large industry groups---by drawing attention to challenging and important open problems that can be explored on a university budget.
In recent years, efforts aimed at widening participation in LM research often take different avenues from the one proposed here, including aggregation of distributed computation power \citep{diskin2021distributed}, reliance on public computing infrastructure \citep{scao-etal-2022-bloom}, aggregation of expertise, data and stepwise contributions \citep{don2022cold,raffel2023building} and modularity \citep{pfeiffer2023modular}. Such a line of pretraining research proposes to keep costs large but to distribute them across funding sources through many contributing factors. 

Other works on decentralizing computation \citep{diskin2021distributed, li2022branch, lialin2023stack} or model recycling works generally take existing models and build upon them, proposing a single adaptation finetuning \citep{choshen2022start}, a single knowledge edit \citep{DeCao2021EditingFK}, combining several models \citep{Yadav2023ResolvingIW}, or iterative approaches showing that stacking such improvements can continually improve models \citep{don2022cold}. Recently, a framework for doing so was also released \citep{git-theta}.
One can see the BabyLM challenge in this context as a suggestion to persist in using a centralized approach to pretraining, but making it tractable, by reducing the cost through increased focus on tractable research questions.\looseness=-1

\section{Guidelines and Timeline} \label{sec:instructions} 


\paragraph{Tracks.}
Submissions to BabyLM had to conform to one of three sets of guidelines, which we term \defn{tracks}. 
In this section, we describe each competition track; for specific details about wording, see the original Call for Papers \citep{cfp}.
The three tracks for the BabyLM challenges were \defn{\strict}, \defn{\smallstr}, and \defn{\loose}. 
Participants in all tracks were allowed a constant number of English-language training tokens (100 million in \strict and \loose and 10 million in \smallstr) to be used in total for all software used in the pipeline. This data was released by the organizing committee and is described, in detail, in \Cref{sec:data}. \loose track submissions were encouraged to train on data beyond just the linguistic text data provided through the shared task (e.g., speech audio signal, code, music, or visual input).
The \loose track also permitted the use of expert-annotated data, but any language data used to train the LM or auxiliary models counted towards the 100M word budget. 
Thus, for example, a \loose track submission could train a parser on the Penn Treebank \citep{marcus1993buildinga} and self-train to parse the pretraining corpus, as long as the number of words in the Penn Treebank plus the pretraining corpus total less than 100M.%
\footnote{
In our initial announcement, external software trained on linguistic input or expert annotations not included in our corpus---including taggers, parsers, tokenizers, or models were \emph{not} allowed. 
However, numerous questions from participants prompted an announcement in April 2023 that we were modifying the rules of the \loose track to allow such methods.
We made this decision because we determined that the interests of the community were better served by emphasizing creativity and discovery in the \loose track.
Text generated by a language model that was trained only on a BabyLM corpus was not counted towards the 100M word budget, nor was data bootstrapped by such models.
}

In general, seeing the same data twice (e.g., across different epochs) did not count as seeing more text. While it is unlikely that humans process data iteratively in a manner similar to epoch-based training, there is evidence that humans do repeat some of the information they process (e.g., in memory replay, \citealp{carr2011hippocampal}). Furthermore, epochs are very useful for gradient-based methods.

Finally, participants across all tracks were encouraged to submit models and papers even if their work did not fit into any of the three tracks. As the goal of the shared task is to advance efficient and cognitively plausible LM training, we did not want to curtail participant creativity. While submissions using external linguistic data did not qualify to win any of the tracks, they still qualified to be presented in the competition and to be published in the proceedings.\looseness=-1

\paragraph{Community building.}
Given that the BabyLM Challenge aims to encourage research in efficient and cognitively plausible model pretraining, one of our goals was to encourage the formation of a research community with shared interests. Towards that end, we hosted a public messaging forum on Slack and enabled participants to interact with each other and with the task organizers. At the time of paper writing, this forum had over 250 members, including many interested researchers who did not ultimately submit to the challenge. An interactive forum was useful for both establishing a community and building interest; it allowed the community to clarify the track rules, debug the evaluation pipeline, and receive announcements from the organizers. 

\paragraph{Timeline.}
Below, we replicate the timeline from the \href{https://babylm.github.io}{website}.
\vspace{-0.7em}
\begin{itemize}[leftmargin=*]
\setlength\itemsep{-0.3em}
    \item \underline{December 2022:} The BabyLM Challenge is announced at CoNLL 2022, as well as on Twitter and in several mailing lists.
    \item \underline{January 2023:} The pretraining datasets for the \strict and \smallstr tracks were released.
    \item \underline{March 2023:} The initial evaluation pipeline was made public.
    \item \underline{1 June 2023:} Hidden (surprise) evaluations were released and the \dynabench submission portal was opened.
    \item \underline{22 June 2023:} Deadline for model submissions (extended from 15 June 2023).
    \item \underline{1 August 2023:} Deadline for paper submissions.
    \item \underline{6-7 December 2023:} Presentation of the shared task at CoNLL.
\end{itemize}

\begin{table*}[t]
    \centering
    \resizebox{\linewidth}{!}{
    \begin{tabular}{llrrr}
    \toprule
    & & \multicolumn{2}{c}{\textbf{\# Words}} & \\\cmidrule(lr){3-4}
    \bf Dataset & \bf Domain & \bf \smallstr & \bf \strict & \bf Proportion \\
    \midrule
    CHILDES \citep{macwhinney2000childes} & Child-directed speech & 0.44M & 4.21M & 5\% \\
    British National Corpus (BNC),\textsuperscript{1} dialogue portion & Dialogue & 0.86M & 8.16M & 8\% \\
    Children's Book Test \citep{hill-2016-cbt} & Children's books & 0.57M & 5.55M & 6\% \\
    Children's Stories Text Corpus\textsuperscript{2} & Children's books & 0.34M & 3.22M & 3\% \\
    Standardized Project Gutenberg Corpus \citep{gerlach2020standardized} & Written English & 0.99M & 9.46M & 10\% \\
    OpenSubtitles \citep{lison2016opensubtitles} & Movie subtitles & 3.09M & 31.28M & 31\% \\
    QCRI Educational Domain Corpus (QED; \citealp{abdelali-etal-2014-qed}) & Educational video subtitles & 1.04M & 10.24M & 11\% \\
    Wikipedia\textsuperscript{3} & Wikipedia (English) & 0.99M & 10.08M & 10\% \\
    Simple Wikipedia\textsuperscript{4} & Wikipedia (Simple English) & 1.52M & 14.66M & 15\% \\
    Switchboard Dialog Act Corpus \citep{Stolcke-etal:2000} & Dialogue & 0.12M & 1.18M & 1\% \\
    \midrule
    \emph{Total} & -- & 9.96M & 98.04M & 100\% \\
    \bottomrule
    \end{tabular}}
    \caption{The datasets we release for the \strict and \smallstr tracks of the BabyLM Challenge. We present the number of words in the training set of each corpus that we include. \textsuperscript{1}\url{http://www.natcorp.ox.ac.uk}\ \ \ \textsuperscript{2}\url{https://www.kaggle.com/datasets/edenbd/children-stories-text-corpus}\ \ \ \textsuperscript{3}\url{https://dumps.wikimedia.org/enwiki/20221220/}\ \ \ \textsuperscript{4}\url{https://dumps.wikimedia.org/simplewiki/20221201/}}
    \label{tab:data}
\end{table*}

\section{Pretraining Corpus}\label{sec:data} 

We compiled and distributed a pretraining corpus inspired by the input received by children.\footnote{Clicking on the following link will download the dataset (240MB zipped, 700MB unzipped): \href{https://github.com/babylm/babylm.github.io/raw/main/babylm_data.zip}{\url{https://github.com/babylm/babylm.github.io/raw/main/babylm_data.zip}}} Submissions to the \strict~track are required to train exclusively on this corpus. Submissions to the \smallstr track are required to use only a scaled-down version of the dataset, approximately 10\% the size of the \strict-track corpus. 
Two key properties of the dataset---its size and its domain---are controlled in order to make the data more developmentally plausible than typical LM pretraining data.\looseness=-1 

\paragraph{Size: 100M words or less.} The pretraining corpus for the \strict track consists of under 100M words, and the corpus for the \smallstr track is under 10M words. 
Children are exposed to 2M-7M words per year \citep{gilkerson2017mapping}. 
Choosing the beginning of adolescence (age 12) as a cutoff, the dataset should be between 24M-84M words, which we round up to 100M words. 
The 10M word \smallstr dataset corresponds to the amount of input in the first two to five years of development.
By contrast, contemporary widely used LMs such as Llama 2 \citep{Llama2} are trained on trillions of words (Figure \ref{fig:data-scale}). 
Even BERT \citep{devlin-etal-2019-bert}, which is comparatively small by today's standards, was trained on over 3B words, well over the amount of input to a human in an entire lifetime. 
This discrepancy in input volume between LMs and humans is an oft-cited criticism of using these artifacts out-of-the-box as cognitive models \citep[a.o.]{warstadt2022artificial,frank2023bridging}.\looseness=-1

\paragraph{Domain: Mostly transcribed speech.} 
We source the majority ($\approx 56\%$) of the pretraining corpus from transcribed or scripted speech. 
We made this choice because the majority of the input to a hearing child comes from speech (though this proportion decreases with age as consumption of written media increases).
This contrasts with standard LM training corpora, which consist mostly of text that was intended to be read and potentially edited. 
This is particularly significant for studying grammar learning, as some grammatical constructions (such as nominalizations and passives) are far more frequent in writing, while others (such as first- and second-person pronouns) are more frequent in speech \citep{biber1991variation}.\looseness=-1

\paragraph{Domain: Child-directed language.} About 40\% of the data in the pretraining corpus comes from sources either intended for children or appropriate for children, including child-directed speech, children's books, educational videos, and simplified English. Child-directed speech has been used as the sole or primary data source in some previous work aiming to model child language acquisition with LMs \citep{reali2005uncoveringa,perfors2011learnability,pannitto2020recurrent,huebner2021babyberta,yedetore-etal-2023-poor}. We chose to include data from other domains (both child-directed and not) for several reasons. First, fewer than 10M words of transcribed child-directed speech are available, far below our 100M word budget. Second, child-directed speech makes up only part of the input to children. This amount can vary by a factor of 10 or more across cultures and socio-economic groups \citep{cristia2019child}. The estimate on which we base the 100M word budget \citep{gilkerson2017mapping} counts \emph{all} speech in the child's environment including overheard speech.

\subsection{Contents}

The contents of the BabyLM pretraining dataset are summarized in Table \ref{tab:data}. Descriptions of each data source are provided in \Cref{app:data_sources}.

\subsection{Preprocessing}

We release \strict and \smallstr train, development, and test splits of each of the ten data sources, split approximately 83.3\%/8.3\%/8.3\%. The 10M word \smallstr  training set is sampled randomly from the \strict training set. After any preprocessing, we downsample and split each source by randomly sampling chunks of 2000 lines or longer. The code and instructions for downloading and preprocessing the raw data are publicly available.\footnote{\href{https://github.com/babylm/babylm_data_preprocessing}{\url{https://github.com/babylm/babylm_data_preprocessing}}.}

We perform minimal preprocessing in terms of filtering and reformatting text. 
Notably, we generally preserve newlines in the original texts, meaning newlines do not consistently delimit documents, paragraphs, or sentences, as in some pretraining datasets. 
We use WikiExtractor \citep{Wikiextractor2015} to extract text from the \texttt{xml} Simple English Wikipedia dump dated 2022-12-01. 
We perform additional preprocessing on Simple English Wikipedia to remove \texttt{<doc>} tags. 
We select the spoken subset of the BNC by selecting only lines from the \texttt{xml} containing the \texttt{<stext>} tag and extracting only the text from the \texttt{xml}. 
We use code by \citet{gerlach2020standardized} to download and preprocess data from Project Gutenberg, which we additionally filter to contain only English texts by authors born after 1850. 
The OpenSubtitles and Wikipedia portions of the pretraining corpus were shared with us in preprocessed form, having had duplicate documents removed from OpenSubtitles and preprocessing steps performed to Wikipedia similar to our Simple English Wikipedia procedure.\footnote{We thank Haau-Sing Li for allowing us to use this preprocessed data.} 
We use regular expressions to remove speaker and dialog act annotations from the Switchboard Dialog Act Corpus.
We perform no preprocessing on the remaining datasets.\looseness=-1

\section{Evaluation} \label{sec:evaluation} 
To evaluate submissions, participants were asked to upload their model predictions to \dynabench, which is an online platform for dynamic data collection and model benchmarking.\footnote{\url{https://dynabench.org/}} Multiple submissions to the \dynabench platform were allowed, but at most one candidate was allowed to be chosen as a competitor from each team.

\subsection{Evaluation Tasks}

The goal of the evaluation pipeline is to assess the extent to which submitted models have learned the latent syntactic and semantic structure of their pretraining language. 
To evaluate the grammatical abilities of LMs, we use BLiMP \citep{warstadt-etal-2020-blimp-benchmark-}. BLiMP consists of tasks that evaluate the ability of language models to behave in a manner consistent with the structure of English. Each example consists of a minimal pair of sentences, where one sentence is acceptable and the other is unacceptable (differing as minimally as possible from the acceptable sentence otherwise); a model is correct on a given example if it assigns higher probability to the correct sentence in the minimal pair. We also release a supplement to the BLiMP tasks, which tests for phenomena not captured by BLiMP (see \S\ref{sec:hidden_eval}).

To assess the abilities of LMs on more typical downstream NLP tasks, we evaluate on a mixture of tasks from a subsample of (Super)GLUE, which consists of text classification tasks. 
We include a variety of task types, including paraphrase detection (MRPC, QQP), sentiment classification (SST-2), natural language inference (MNLI, QNLI, RTE), question answering (BoolQ, MultiRC), acceptability judgments (CoLA), and commonsense reasoning (WSC).\looseness=-1


\subsubsection{Hidden Tasks}\label{sec:hidden_eval} 
Two weeks before the results deadline, we released three hidden evaluation tasks: the Mixed Signals Generalization Set (MSGS), a supplement to BLiMP, and an age-of-acquisition (AoA) prediction task. MSGS and the BLiMP supplement were mandatory; AoA prediction was provided as an additional analysis point for participants in writing their papers. The motivation for using these hidden tasks was to prevent our evaluations from rewarding submissions that overfit to the BLiMP and (Super)GLUE tasks.

The BLiMP supplement includes five test suites consisting of BLiMP-style minimal pairs that cover areas of linguistic knowledge not tested by BLiMP---namely, dialogue and questions. The test suites are semi-automatically generated using manually filled templates. As with BLiMP, models are evaluated on the supplement in a zero-shot manner, by comparing the probabilities of the sequences in a minimal pair, under the assumption that the acceptable sequence will be more probable than its unacceptable counterpart.

\paragraph{\testsuite{Hypernyms}.} We evaluate LMs' knowledge of lexical entailment, i.e., hypernym--hyponym relationships. This task bears similarity to natural language inference \citep{dagan2006pascala,bowman2015large, williams-etal-2018-broad}, but we instead measure whether models assign a higher likelihood to valid statements of entailment compared to minimally differing invalid statements. The evaluation data is designed around manually written triples consisting of $\langle\hypernym, \base, \hyponym\rangle$---for example, $\langle\emph{plant}, \emph{herb}, \emph{basil}\rangle$. We also specify an \other\ noun (for example, \emph{flower}) which shares the \hypernym\ but not the \hyponym\ with the \base\ noun. From these nouns, plus a set of manually written contexts, we generate six types of minimal pairs, shown in Table~\ref{tab:hypernym} in \Cref{app:blimp_supplement}. Additionally, we randomly vary the text used to convey entailment, e.g., \textit{If p then q}, \textit{If p that means q}, \textit{p therefore q}, etc.\looseness=-1

\paragraph{\testsuite{Subject--Auxiliary Inversion}.}~The subject--auxiliary inversion rule applies in question formation in English (e.g., relating \textit{Logan will go.} to \textit{Will Logan go?}). This task has been used to evaluate language models' syntactic abilities and preferences \citep[e.g.,][]{mccoy-etal-2020-syntax,mueller-etal-2022-coloring,yedetore-etal-2023-poor,mueller-linzen-2023-plant}. Our test data was created by \citet[Ch. 6]{warstadt2022thesis}, where it is described in more detail.\looseness=-1

\paragraph{\testsuite{Turn-Taking}.}
Comprehending dialogue requires tracking the grammatical properties of utterances from multiple speakers. Pronouns such as \emph{I}, \emph{you}, and \emph{she} are indexicals, meaning their interpretation depends on the speaker's context and identity. This test suite evaluates whether LMs can predict which pronoun is appropriate to use when there is a change in speaker. For example, if person A asks person B a question of the form \emph{Can I ...}, person B's response should begin with \emph{You}, not \emph{I}. Our tests include (i) cases where the pronoun is expected to change, and (ii) cases where it is not. We also vary the context length (and therefore the distance between the context pronoun and the target), and whether the context contains a distractor pronoun in an embedded position. Finally, for each example, we randomly select one from a set of formats for indicating the speaker, e.g., \emph{A: ..., B: ...}, or \emph{``...,'' he asked. ``...,'' she said.}, etc. Examples of each format can be found in Table~\ref{tab:turntaking1} in \Cref{app:blimp_supplement}.

\paragraph{\testsuite{Question--Answer Congruence}.}
The syntax of a question constrains the acceptable responses. For example, a congruent answer to a \emph{who}-question must be an animate noun (or contain one in a suitable context). This test suite evaluates whether LMs assign a higher likelihood to congruent answers compared to incongruent ones, and therefore learn the cross-sentential dependency between a \emph{wh}-word and an answer. In addition to a set of \testsuite{easy} test cases, we construct a set of adversarial \testsuite{tricky} test cases where there is a highly salient distractor answer that is not congruent with the \emph{wh}-word. We randomly vary whether the answer appears as a fragment or in a complete sentence as well as the format for indicating the speaker. See Table~\ref{tab:turntaking2} in \Cref{app:blimp_supplement} for examples.\looseness=-1

\paragraph{Mixed Signals Generalization Set.} The Mixed Signals Generalization Set (MSGS; \citealp{warstadt-etal-2020-learning}) is a text classification task that evaluates the inductive biases of language models. For a MSGS subtask, models are finetuned on an ambiguous training set where the labels are consistent with both a syntactic generalization and a surface generalization, and then evaluated on examples that disambiguate which generalization the model converged on (if any).\footnote{
For example, one of the subtasks tests which of the following two generalizations the model's inductive bias favors:
whether the word ``the'' is present (the surface generalization), or whether the sentence contains an adjective (the syntactic generalization).
Thus, training examples will include only ambiguous labeled pairs where these two properties are both perfectly correlated with each other and with the binary labels, such as \texttt{(The big dog barked, 1)} and \texttt{(A dog barked, 0)}.
At test time, the model must classify held-out sentences where the features are anti-correlated, such as \texttt{A big dog barked} and \texttt{The dog barked}.
If the model predicts labels \texttt{1} and \texttt{0} respectively for these and other analogous examples, we infer that it classifies examples based on the linguistic feature, while if it predicts \texttt{0} and \texttt{1} respectively, it adopted the surface generalization.
}

Ideally, models would be more sensitive to linguistic features than surface features, as a systematic preference for abstract linguistic properties allows models to generalize more robustly to unseen structures.
The metric for MSGS is the Matthews correlation coefficient between the model's predictions and the labels according to the linguistic generalization on the test set.
A coefficient of 1 corresponds to a systematic linguistic generalization, and -1 to a systematic surface generalization.
Indeed, \citet{warstadt-etal-2020-learning-} find that linguistic bias increases with the volume of pretraining data, and that models with RoBERTa-like architectures require more than a billion words of pretraining data to achieve an overall linguistic bias (i.e., a score greater than 0).

\paragraph{Age-of-acquisition Prediction.} Optionally, participants could evaluate on the age of acquisition (AoA) prediction task of \citet{portelance2023predicting}. When humans are learning language, they tend to acquire certain words at specific ages; the age of acquisition of a word refers to the age at which humans acquire that word. The AoA prediction task compares LMs' word surprisals with children's AoA of the same words. A language model's average surprisals are converted into AoA predictions, and these are then compared to the actual average AoA (in months) of those words. Models achieving lower mean absolute deviation between the actual age and predicted age are said to perform better on the task.\footnote{It is not clear whether optimizing LM performance on this task necessarily leads to better language models. It is possible instead that LMs could have a different pattern of surprisals than humans while learning particular linguistic concepts more or less efficiently than humans. Thus, this task should be used more as a measure of how well LMs align with humans---and thus, as a measure of their usefulness as cognitive models of language acquisition and processing---rather than as a measure of quality or performance.} While we did not require participants to submit these scores as part of their predictions, we provided code to make evaluation on this task simple, such that they could include this score as an additional analysis point in their paper submissions. 7 teams (22.6\%) evaluated on the AoA prediction task; see Appendix~\ref{app:aoa_eval} for results and discussion.

\subsection{Evaluation Pipeline}\label{sec:eval-pipeline}
The organizers provided code to unify the evaluation setup across submissions. This was released as a public repository on GitHub.\footnote{\url{https://github.com/babylm/evaluation-pipeline}} The evaluation pipeline supports models implemented in HuggingFace, though we did not restrict the model submissions to HuggingFace-based models.\footnote{Upon release of the evaluation pipeline, we announced that we would provide support as needed to teams training LMs not based in HuggingFace.} For model and result submissions, users were required to (i) upload a link to their model (on any file-hosting service), and (ii) provide model predictions for each example of each task (via \dynabench); we provided a template specifying the format of the predictions file.

\paragraph{Data preprocessing.} NLP tasks in our evaluation pipeline often contained vocabulary that is not contained in the BabyLM pretraining corpora. To address this mismatch, we filtered each task according to its lexical content: if an example contained any words that appear less than twice in the \smallstr training corpus, we filtered the example out. Otherwise, each dataset is presented in its original format. See Table~\ref{tab:eval-data-size} in Appendix~\ref{app:eval-data-size} for details on the size of the filtered datasets.

\subsubsection{Evaluation Paradigms}
\paragraph{Zero-shot evaluation.} For zero-shot tasks---BLiMP and the BLiMP supplement---we modify the BigScience fork of the \texttt{lm-eval-harness} repository, originally by EleutherAI \citep{eval-harness}. This provides functionality for scoring autoregressive decoder-only LMs and encoder-decoder LMs. For encoder-only LMs, we modify the repository to support masked language model scoring as described in \citet{salazar-etal-2020-masked}.\footnote{We use the implementation of \citet{misra2022minicons} in the \texttt{minicons} library.}

\paragraph{Finetuning.} We first attempted zero-shot learning and few-shot in-context learning for (Super)GLUE and MSGS tasks. However, this often resulted in random-chance accuracies from each of our baselines; we, therefore employ finetuning.\footnote{finetuning technically adds to the training set size. We consider this acceptable, as finetuning on a single GLUE or MSGS task does not meaningfully add to the domain-general linguistic abilities of language models. The LM is finetuned separately for each task, so we still see this as an evaluation of the LM's abilities in itself (albeit more confounded than the zero-shot evaluations).} For tasks requiring finetuning---(Super)GLUE \citep{wang-etal-2018-glue,wang-etal-2019-superglue} and MSGS \citep{warstadt-etal-2020-learning}---we base our scripts on HuggingFace's example finetuning scripts for text classification.\footnote{\url{https://github.com/huggingface/transformers/blob/211f93aab95d1c683494e61c3cf8ff10e1f5d6b7/examples/pytorch/text-classification/run_glue.py}} We modified the script to support encoder-decoder models, and to work for a wider variety of tasks. We provide a default set of hyperparameters that we found to work well across our baseline models, though participants were allowed to freely modify hyperparameters.\looseness=-1

\subsection{\dynabench Leaderboard}
\dynabench is an open-source platform for dynamic dataset creation, model evaluation, and leaderboard hosting \citep{kiela-etal-2021-dynabench}. In addition to open-sourcing datasets---including adversarial and human-in-the-loop datasets \citep{nie-etal-2020-adversarial, bartolo-etal-2021-improving, potts-etal-2021-dynasent, sheng-etal-2021-human, vidgen-etal-2021-learning, kirk-etal-2022-hatemoji}---\dynabench has offered leaderboard support for several community challenges in the past \citep{wenzek-etal-2021-findings,dadc-2022-dynamic, dataperf}. Given that we desire a dynamic leaderboard that allows for submissions even after the end of the challenge, this platform was well-suited to the BabyLM Challenge. 
All model submissions to the challenge were submitted via the \dynabench platform, to the respective leaderboards for the \strict,\footnote{\href{dynabench.org/tasks/baby_strict}{\url{https://dynabench.org/tasks/baby_strict}}} \smallstr,\footnote{\href{dynabench.org/tasks/baby_strict_small}{\url{https://dynabench.org/tasks/baby_strict_small}}} and \loose\footnote{\href{dynabench.org/tasks/baby_loose}{\url{https://dynabench.org/tasks/baby_loose}}} tracks. 

Each leaderboard presents aggregate scores across all tasks, which can be interactively broken down into more fine-grained scores per task and per subtask. 
To compute the aggregate score, we weigh BLiMP and the BLiMP-supplement together at 50\% (all subtasks weighted equally), (Super)GLUE at 30\%, and MSGS at 20\%.
This weighting scheme was arrived at heuristically, though we did observe that the winners for each track were stable across a wide range of reasonable weightings.
Dynabench allows users to specify a custom task weighting to compute an alternative aggregate score.
The leaderboard for the BabyLM challenge will continue to accept submissions indefinitely.

\subsection{Baselines and Skylines} \label{sec:baselines}

\begin{figure}[t]
    \centering
    \includegraphics[width=\columnwidth]{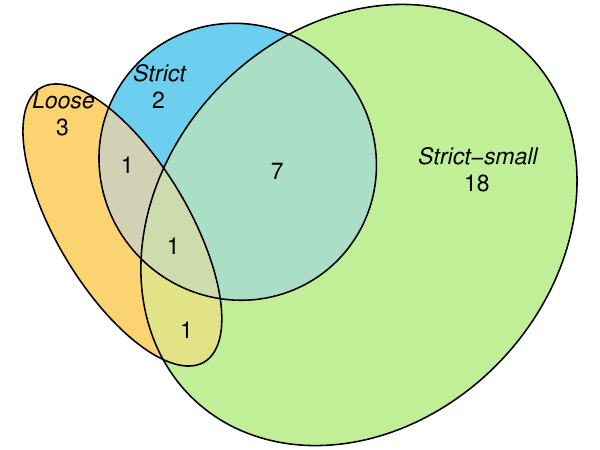}
    \caption{Number of participants who submitted to each track, with multiple submissions counted once.\todo{Replace with final data}}
    \label{fig:submission_counts_euler}
\end{figure}

\paragraph{Baselines.} 
To provide simple baselines for our evaluation tasks, we train multiple models on the data released for \smallstr and \strict tracks and evaluate them on the evaluation tasks.
Three baseline models are provided: OPT-125M, RoBERTa-base, and T5-base.
These models use the same objective function and network architecture corresponding to their original papers (OPT;~\citealp{zhang-etal-2022-opt}, RoBERTa;~\citealp{liu-etal-2019-roberta}, T5;~\citealp{raffel-etal-2020-t5}). 
The network architecture of these models covers both encoder-decoder (T5-base and RoBERTa-base) and decoder-only (OPT-125M) architectures. 
Their objective functions include next-token prediction (OPT-125M), masked-token prediction (RoBERTa-base), and sequence-to-sequence (T5-base) matching losses.
The baseline models are trained using a fixed context length of 128, a constant learning rate of 1e-4, a linear learning-rate warmup from 0 in the first 5000 steps, a batch size of 128, and AdamW~\cite{AdamW} as the optimizer.
They are trained for 20 epochs on the data, where each epoch randomly and independently shuffles the whole dataset.
Although most of these hyperparameters are loosely inspired by \citeauthor{huebner2021babyberta}, we expect that the specific choices on them can be further improved and leave these potential improvements as possible topics for submissions.
We find that our baseline models achieve reasonable performance on the evaluation tasks, with clear improvement from more data from \smallstr to \strict track and notable gap towards their counterparts pretrained on much larger datasets.\looseness=-1

\paragraph{Skylines.} To get an approximation of how well larger models could, in principle, perform in our task and setting, we ran Llama 2 70B \citep{Llama2} and the fully trained RoBERTa-base model through our evaluation pipeline. 
This is meant to provide a comparison point to the state of the art in 2023, as the Llama 2 model is pretrained on much more data (2T tokens) than the challenge allows, and it has far more parameters than we expect to find in submissions.
We evaluate Llama 2 on (Super)GLUE using in-context learning, but it is fully finetuned on MSGS.
BabyLM submissions that approach these scores can be considered to have greater sample efficiency than the skyline models, and may therefore provide stronger starting points for future research in sample-efficient NLP.\looseness=-1

\begin{figure}[t]
    \centering
    \includegraphics[width=\columnwidth]{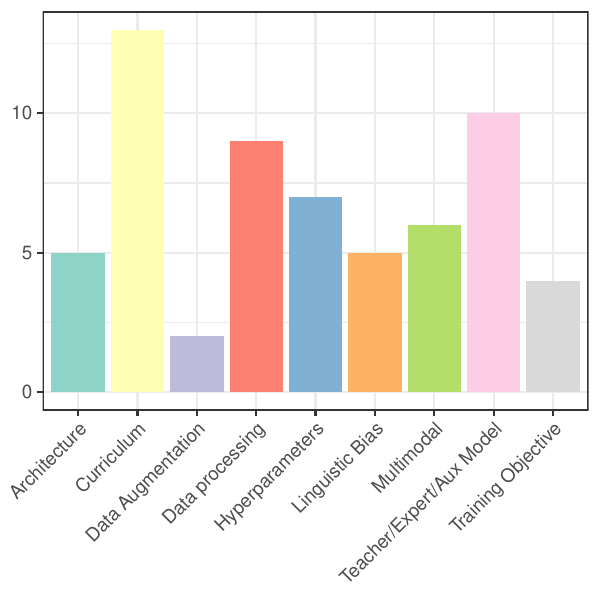}
    \caption{Total number of submitted models that used each of the nine approaches in our typology. We count at most one submitted model per participant per track.}
    \label{fig:approach_count}
\end{figure}

\section{Submissions Summary} \label{sec:submissions} 


\begin{table*}[tbhp]
    \centering
    \resizebox{\linewidth}{!}{
    \begin{tabular}{clrrrrr}
    \toprule 
        & \textbf{Model} & BLiMP & GLUE & MSGS & BLiMP-Supp. & \emph{Aggregated}\\
    \hline
         & Llama 2 & 0.84 & 0.84 & 0.26 & 0.75 & 0.71 \\
         & RoBERTa-Base & 0.87 & 0.79 & 0.24 & 0.76 & 0.70 \\
    \hline
        \parbox[t]{2mm}{\multirow{4}{*}{\rotatebox[origin=c]{90}{\small{Strict}}}} & ELC-BERT~\cite{babylm2023notall} & 0.85 & 0.78 & \underline{\textbf{0.47}} & \underline{\textbf{0.77}} & \underline{\textbf{0.74}} \\
         & BootBERT~\cite{babylm2023meanberts} & \underline{\textbf{0.86}} & \underline{\textbf{0.79}} & 0.28 & 0.72 & 0.70 \\
         & McGill-BERT~\cite{babylm2023mcgill} & 0.84 & 0.72 & 0.25 & 0.71 & 0.67 \\ 
         & \emph{Best Baseline (OPT-125M)} & 0.75 & 0.70 & 0.13 & 0.68 & 0.60 \\
    \hline
        \parbox[t]{2mm}{\multirow{4}{*}{\rotatebox[origin=c]{90}{\small{Strict-Small}}}} & ELC-BERT~\cite{babylm2023notall} & \underline{\textbf{0.80}} & \underline{\textbf{0.74}} & \underline{\textbf{0.29}} & 0.67 & \underline{\textbf{0.66}} \\
         & MLSM~\cite{babylm2023bettertogether} & 0.79 & 0.71 & 0.17 & 0.57 & 0.61 \\
         & McGill-BERT~\cite{babylm2023mcgill} & 0.75 & 0.70 & 0.13 & \underline{\textbf{0.68}} & 0.60 \\
         & \emph{Best Baseline (OPT-125M)} & 0.63 & 0.62 & 0.10 & 0.53 & 0.50 \\
    \hline
        \parbox[t]{2mm}{\multirow{3}{*}{\rotatebox[origin=c]{90}{\small{Loose}}}} & Contextualizer~\cite{babylm2023contextualizer} & \underline{\textbf{0.86}} & \underline{\textbf{0.73}} & \underline{\textbf{0.58}} & 0.63 & \underline{\textbf{0.73}} \\
         & McGill-BERT~\cite{babylm2023mcgill} & 0.80 & 0.68 & -0.02 & 0.57 & 0.57 \\
         & BabyStories~\cite{babylm2023babystories} & 0.78 & 0.61 & 0.03 & \underline{\textbf{0.65}} & 0.56 \\
    \bottomrule
    \end{tabular}}
    \caption{Top 3 systems for each track, as well as the baseline model with the highest aggregate score. We also show ``skyline'' models: RoBERTa-base and Llama 2 trained on their full pre-training corpora. Each task score is simply the mean score across each of its subtasks. The aggregate score is a weighted average of each task. We \underline{\textbf{bold}} the highest-scoring system for each task within each track.}
    \label{tab:result}
\end{table*}

\begin{table}[]
    \centering
    \begin{tabular}{lll}
    \toprule
        & \bf \# Models & \bf \# Participants\\\midrule
        \loose & 20 & 8 \\
        \smallstr & 118 & 29 \\
        \strict & 24 & 11 \\
        \midrule
        \it total & \it 162 & \it 31 \\\bottomrule
    \end{tabular}
    \caption{Total number of models and participants per track. Participants who submitted to multiple tracks are counted once in the total. \todo{Replace with final data}}
    \label{tab:submission_counts}
\end{table}

We received 31 papers and 162 models in total. Table \ref{tab:submission_counts} shows the submission counts for each track. Some participants submitted to multiple tracks; we show data for unique participants in Figure \ref{fig:submission_counts_euler}. 

We found that many submissions focused their efforts on similar techniques. To better quantify this, we devised a typology of the nine most common approaches and assigned each submitted model one or more labels. Figure \ref{fig:approach_count} shows the number of submissions employing each approach. \S\ref{subsec:common_methods} provides more detailed descriptions of each approach, as well as results indicating which ones were most effective.

All participants are affiliated with universities or independent research institutions. Participants' home institutions are located in 16 different countries. The number of participants by country is as follows (multinational participants are counted more than once): US (9), Germany (5), Netherlands (3), UK (4), Canada (2), Norway (2), Austria (1), Denmark (1), France (1), Hungary (1), Israel (1), Japan (1), Norway (1), Switzerland (1), Turkey (1).\looseness=-1

The official leaderboard is available on  \dynabench.\footnote{\url{https://dynabench.org/babylm}}
With the consent of participants, we release links to submitted models, their complete predictions for the evaluation tasks, their scores for each task and subtask, and metadata about each submission at the BabyLM's GitHub at \url{https://github.com/babylm/submissions2023}.
We provide a summary of each submission in Appendix~\ref{app:summaries}.\looseness=-1

\section{Results \& Analysis} \label{sec:results} 

\begin{figure*}
    \centering
    \includegraphics[width = \linewidth]{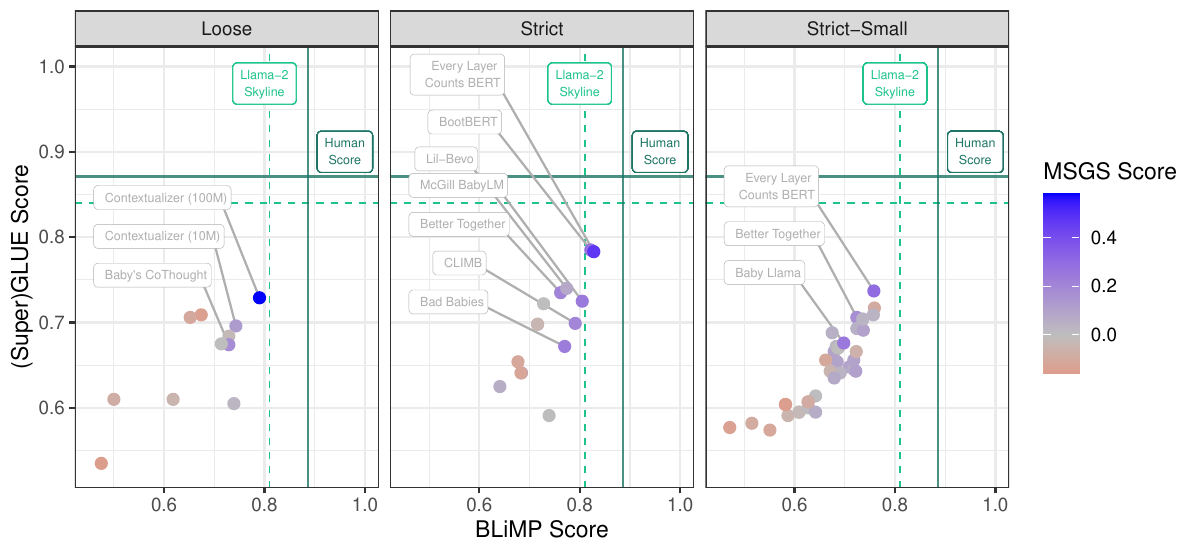}
    \caption{\textbf{Summary of BabyLM Submission Results:} Each point represents an official model submission. Scores are broken down into performance on BLiMP ($x$-axis), GLUE ($y$-axis) and MSGS (color). Submissions that achieve an aggregate score above $0.6$ are labeled in gray. Green dashed lines show Llama 2 skyline performance, and green solid lines show the human performance ceiling.}
    \label{fig:overall-results}
\end{figure*}

\subsection{Overall Results \& Track Winners}

The results from all submissions are shown in \Cref{fig:overall-results}, with the scores of the top-performing models in each track detailed in \Cref{tab:result}. In the figure, dashed green lines show the performance of the Llama 2 skyline. Solid green lines show human performance on GLUE reported in \citet{nangia2019human}, and human performance on BLiMP as reported by \citet{warstadt-etal-2020-blimp-benchmark-}.\looseness=-1

Before discussing the winning systems in each track, we note a few high-level takeaways from these results. The strongest results were achieved by models in the \strict~track. Given the \strict~track's larger training corpus relative to the \smallstr corpus, it is not surprising that these models could outperform those in the \smallstr~track. However, there are two interesting trends: First, \strict models did not outperform those in \smallstr by a large amount, even though the size of training data was an order-of-magnitude larger. For example, there are only two models in the \strict track that achieve higher GLUE scores than the best-performing \smallstr model. Second, models in the \loose track tended to perform worse in the aggregate than those in the \smallstr track, even though they potentially had access to additional (non-linguistic) data. One conclusion we can draw from this is that learning from multiple modalities of data presents a challenge in its own right, and that current model architectures are not optimized to efficiently utilize multiple types of inputs during training.

The other important high-level takeaway is that many BabyLM models are very close to the Llama 2 skyline, and to achieving human-level performance on BLiMP and GLUE (i.e., they are near the green lines in \Cref{fig:overall-results}). Strong performance could be expected in the case of (Super)GLUE, where models were finetuned with additional data, but we note that even for BLiMP, the top-performing model is only about $3\%$ shy of human performance. Note that prior to the start of the challenge, we explored the possibility of measuring zero-shot performance on (Super)GLUE test sets, and found zero-shot performance to be at or below chance for our baselines. This fact, as well as the consideration that GLUE has been traditionally evaluated using finetuning, leads us to select finetuning evaluations for the (Super)GLUE benchmark(s).

Given that successful training on developmentally plausible corpora could have ramifications for cognitive and linguistic theories of learnability \citep{wilcox2023using, warstadt2022artificial}, these results point to two important takeaways: (1) Human-level results have not been achieved \emph{yet}. However, (2) given the strong performance of the top-scoring models, human-level results appear likely to be achieved very soon, possibly within the next few years. Of course, one possible concern is the following: current models may not be close to human-level performance; rather, current performance metrics, like BLiMP, might not accurately measure human-level linguistic competence. We are sympathetic to such concerns, but we also note that BLiMP, and other related syntactic benchmarks such as those presented in \citet{marvin-linzen-2018-targeted} and \citet{gauthier-etal-2020-syntaxgym}, were specifically designed to mimic the types of tests invented by linguists and cognitive scientists to reveal syntactic competence---i.e., they are all based on minimal pair sentences. Thus, while it is imperative to continue building more comprehensive and larger datasets, we believe it is fair to say that the close-to-human scores observed in the BabyLM challenge on BLiMP reflect genuine grammatical generalizations learned by the models.


\subsection{Winning Submissions}\label{sec:winning-submissions}
Below, we discuss the winning submissions from each track in greater detail. We also mention the winners of our ``Most Interesting Paper'' awards and provide a brief justification for each.

\paragraph{\strict track.} The winner of the \strict track is ELC-BERT submitted by \citet{babylm2023notall}. 
This model, as well as the runner-up submission Boot-BERT
\citep{babylm2023meanberts}, used as their starting point the LTG-BERT architecture from \citet{samuel2023trained}.
Although these submissions make additional incremental improvements to the LTG-BERT training regime, their own baselines suggest that the backbone architecture plays a large role in the submissions' successes.
LTG-BERT's main contribution is a synthesis of several optimizations to the Transformer architecture, namely:
(1) additional layer normalization, following \citep{shleifer2022normformer};
(2) GEGLU feed-forward modules \citep{shazeer2020glu};
(3) disentangled attention following DeBERTa \citep{he2021deberta}; and
(4) scaled weight initialization following \cite{nguyen-salazar-2019-transformers}.
ELC-BERT modifies this backbone such that the input to each layer is a weighted sum of the outputs of all previous layers. 
Another notable property of LTG-BERT is that all models with this architecture so far have been trained for a large number of epochs.
\citet{babylm2023notall} train models for over 450 epochs for their \strict submission, and over 2000 epochs for their \smallstr submission.
LTG-BERT models performed exceptionally well on our set of evaluations, outperforming not only every other submission to the shared task but also the Llama 2 and RoBERTa-Base skylines on overall score and on all test suites except for (Super)GLUE (Table \ref{tab:result}).
The second runner-up for this track was McGill-BERT \citep{babylm2023mcgill}.

\begin{figure*}
    \centering
    \includegraphics[width = 0.95\linewidth]{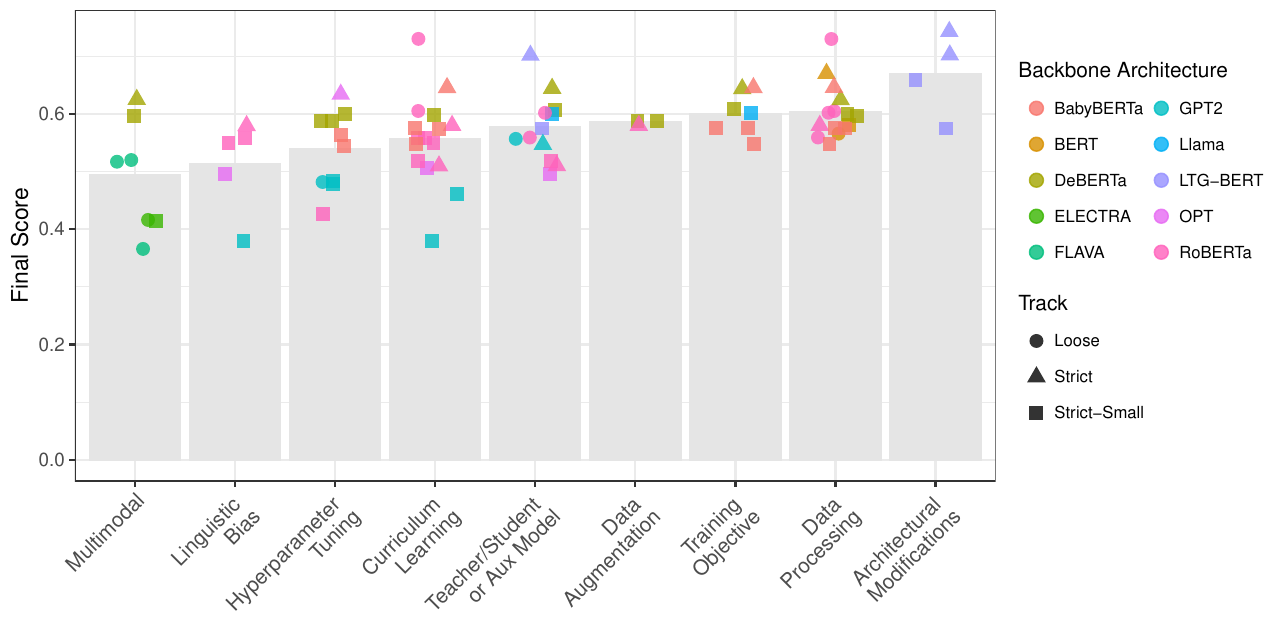}
    \caption{\textbf{Effect of Training Strategy and Backbone Architecture:} Each point represents a submission. Some submissions may appear more than once if they use multiple strategies. Shapes show the challenge track to which the model was submitted. Colors show the backbone architecture on which the model is based. Gray bars show within-category aggregates.}
    \label{fig:method-results}
    \vspace{-15pt}
\end{figure*}

\paragraph{\smallstr track.} The winner of the \smallstr track is, again, ELC-BERT \citep{babylm2023notall}. This double-win demonstrates that the model's architectural choices work well with multiple scales of pretraining data. The runners-up were MLSM \citep{babylm2023bettertogether} and McGill-BERT \citep{babylm2023mcgill}.

\paragraph{\loose track.} The winner of the \loose track is the Contextualizer model of \citet{babylm2023contextualizer}, which used a data processing scheme in which extra training samples are created by combining chunks of texts from different contexts.
Repeating this process 40 times for each chunk gives a dataset that has as many training samples as 4B word dataset, but based on a dataset of only 100M words.
This augmentation technique outperforms training for 40 epochs using the same training samples.
Runners-up for this track were McGill-BERT \citep{babylm2023mcgill} and the BabyStories model of \citet{babylm2023babystories}.\looseness=-1

\subsection{Most interesting paper awards.} These awards are given to papers that go beyond achieving high scores on a leaderboard, and instead demonstrate contributions to the shared task based on interesting analyses, useful negative results, creative modeling choices, or a combination thereof. We awarded two most interesting paper awards in two different categories.\looseness=-1

\paragraph{Outstanding evaluation.} The most interesting paper award for outstanding evaluation was given to ``Large GPT-like Models are Bad Babies: A Closer Look at the Relationship between Linguistic Competence and Psycholinguistic Measures'' \citep{babylm2023gptlike}. This work goes beyond the BabyLM evaluation tasks: the authors use measures of human cognitive processing effort and linguistic competence and additionally correlate these with BabyLM task performance. Their work assesses BabyLM submissions as models of human language processing, thus contributing to our understanding of how to better train cognitive models.

\paragraph{Compelling negative results.}
The most interesting paper award for compelling negative results was given to ``CLIMB---Curriculum Learning for Infant-inspired Model Building'' \citep{babylm2023climb}. This work proposes a typology of common curriculum learning approaches and performs a thorough and principled evaluation exploring this design space. Although they find that none of the tested approaches leads to widespread improvements across the evaluation tasks, the exhaustiveness of this search and the careful controls and baselines in the study make this negative result a valuable contribution.


\subsection{Common Methods}\label{subsec:common_methods}

\begin{figure*}
    \centering
    \includegraphics[width = 0.95\linewidth]{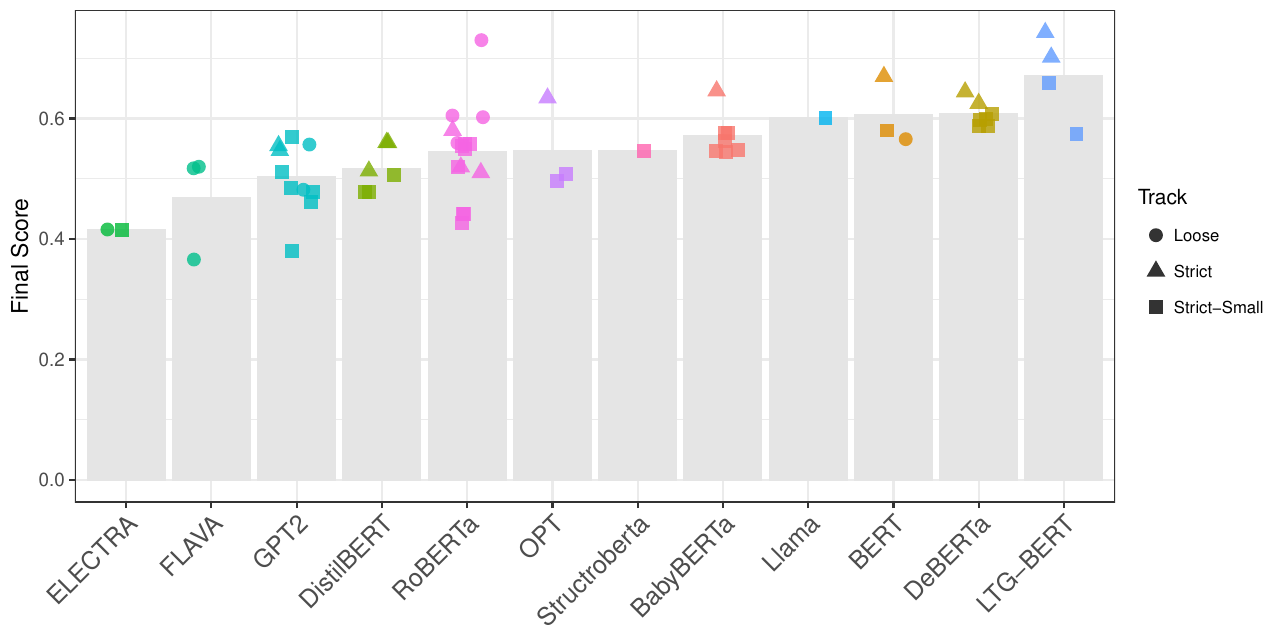}
    \caption{\textbf{Effect of Backbone Architecture:} Each point represents a submission. Shape indicates the challenge track. Gray bars show within-category aggregates.}
    \label{fig:backbone-results}
    \vspace{-15pt}
\end{figure*}

One of the main objectives of the BabyLM Challenge is to compare and contrast methodological choices for sample-efficient pretraining. To do so, we hand-coded each submission based on the method(s) it employs. Figure \ref{fig:approach_count} shows the number of submissions using each approach, and we visualize the performance of different methods in \Cref{fig:method-results}. We also present a similar figure separated by the underlying architecture (\Cref{fig:backbone-results}). Each of these approaches is discussed in further detail below. We highlight two high-level takeaways to start: First, curriculum learning, which was the most popular approach, did not tend to produce high scores (although one curriculum learning model did perform well). Second, the highest-performing models were ones that made architectural modifications---namely, those based on the LTG-BERT architecture.\looseness=-1

\paragraph{Curriculum learning.} This approach entails sorting training steps with respect to some complexity metric(s). This was the most popular approach, with 13 teams (41.9\%) attempting some variant of curriculum learning.
The majority of these attempts did not produce consistent improvements across the BabyLM evaluation tasks. 
However, they did explore a large space of possible curricula, for example:
ranking sentences by 
surprisal \citep{babylm2023cantrain,babylm2023surporc}, 
lexical frequency \citep{babylm2023gpt2opt,babylm2023climb}, 
length \citep{babylm2023byterank,babylm2023toomuch}, and
syntactic complexity \citep{babylm2023mmi01,babylm2023conllst,babylm2023gptwee};
sorting entire datasets by difficulty \citep{babylm2023curriculum,babylm2023climb,babylm2023contextualizer};
gradually increasing vocabulary size \citep{babylm2023cogmemlm,babylm2023toomuch};
and gradually increasing the difficulty of the training objective \citep{babylm2023climb}.\looseness=-1

\paragraph{Teacher--student or auxiliary model.} 
Many papers trained their submitted models with the aid of additional models.
According to our rules, this was permissible as long as any auxiliary models were trained on the BabyLM corpus.
Knowledge distillation using auxiliary models was often a successful approach:
\citet{babylm2023meanberts} considered an exponential moving average teacher model \citep{tarvainan-and-valpola-2017-meanteachers}, 
while \citet{babylm2023bettertogether} modeled a latent semantic feature distribution from a teacher model. 
\citet{babylm2023babyllama} performed distillation on an ensemble of features.
Others used auxiliary models to select appropriate training examples for a curriculum \citep{babylm2023cantrain,babylm2023surporc},
or trained a reward model for use in reinforcement learning \citep{babylm2023babystories}.

\paragraph{Data preprocessing.} Many submissions modified the format of the pretraining corpus. 
When controlled comparisons were performed, these preprocessing steps often led to improvements.
In \S\ref{sec:winning-submissions} we discuss the successful Contextualizer method for constructing new training samples.
Other successful approaches used short sequences or individual sentences as training samples, rather than long portions of documents \citep{babylm2023lilbevo,babylm2023mcgill,babylm2023toomuch}.
Among the more unique approaches in this space was Baby's CoThought \citep{babylm2023babycot}, which used an LLM to reformat unrelated sentences from the corpus into coherent paragraphs.\looseness=-1

\paragraph{Hyperparameter tuning and model scaling.} 
This was a relatively common approach. 
Many submissions performed extensive hyperparameter searches, producing hard-won hyperparameters that work well on smaller datasets while preserving features of the dataset. 
While extensive hyperparameter searching can be expensive and challenging when scaling up to full-sized pretraining, in our limited data regime, consistently successful modifications include reducing context length (see ``Data preprocessing'', above), and training for more epochs or long epochs with data augmentation \citep{babylm2023chapgtp,babylm2023llmsdev,babylm2023pennbgu,babylm2023contextualizer,babylm2023meanberts,babylm2023notall}.

However, results are mixed when modifying model size: some participants achieved better results when scaling model sizes up \citep{babylm2023todberta}, while others were able to perform well when using very small models \citep{babylm2023miniminds}. 
More controlled studies using a variety of architectures and datasets are needed to determine whether scaling up or down is a better solution.

\paragraph{Multimodal learning.} Multimodal learning was one of the directions where we expected the most interest and the most submissions; 
however, we received few submissions based on multimodal inputs, and the multimodal submissions did not reliably contribute to higher overall accuracy. 
One submission used music \citep{babylm2023lilbevo}, another used vision and language data \citep{babylm2023acqling}, a third explored text-and-audio \citep{babylm2023whisbert}, and a fourth incorporated text-and-image data and lexical sensorimotor data as part of the embedding process using multiplex networks \citep{stella2017multiplex,ciaglia-2023-multiplex}. 
Music training produced minor improvements on some subtasks, while the vision-and-language system marginally improved over the baselines in the \smallstr track. 
The multiplex network did not produce performance gains, though it did allow the participants to reduce the number of parameters while preserving performance relative to the baselines. 
WhisBERT was reported to be undertrained, making its results difficult to interpret.\looseness=-1





\paragraph{Architecture modifications.} 
The winning submission made architectural modifications: \citet{babylm2023notall} made slight improvements to LTG-BERT (see \S\ref{sec:winning-submissions} for more on this architecture) by taking a weighted sum over the outputs of all previous layers.
\citet{babylm2023implicitstructure} used the relatively novel StructFormer architecture \citep{shen-etal-2021-structformer1}, which encourages tree-structured representations of inputs.

\paragraph{Training objectives.} 
Some submissions trained language models using a mixture of both a language modeling objective and some other objective. 
Knowledge distillation from teacher models (see paragraphed titled ``Teacher--student or auxiliary model'' above) was the most common modification.
\citet{babylm2023climb} simplified the masked language modeling objective by coarse-graining the output classes, with little effect.
\citet{babylm2023lilbevo} achieved improvements on specific BLiMP subtasks by modifying the masking procedure to preferentially mask specific words thought to be relevant to a particular phenomenon tested by BLiMP.\looseness=-1

\paragraph{Linguistic bias.} Some submissions tried to impart human linguistic biases to models. 
Such approaches discussed above include curriculum learning based on linguistically motivated data sorting methods and architectures like StructFormer that encourage hierarchical analyses of inputs.
\citet{babylm2023grammar} also pretrained with token embeddings obtained via grammar induction, and \citet{babylm2023cogmemlm} iteratively updated the vocabulary of the LM based on word simplicity measures (motivated by human age-of-acquisition analyses).

\paragraph{Data augmentation.} 
Arguably, the effective Contextualizer approach \citep{babylm2023contextualizer} is a form of data augmentation (see \S\ref{sec:winning-submissions}).
\citet{babylm2023chapgtp} used regular expressions to generate question-answer pairs given the BabyLM training data.
\citet{babylm2023babystories} used an LLM to generate text merging disparate sentences into cohesive paragraphs.








\section{Future BabyLM Challenges}\label{sec:future}
The first iteration of the BabyLM Challenge yielded many successes, but also some organizational and scientific challenges.
The lessons learned from our findings can improve future iterations of this challenge.

We were surprised that there were significantly more submissions to the \smallstr~track than the other two tracks combined, considering that the \loose~track allows for a much wider variety of methods. However, this is understandable from the perspective of compute: training on \smallstr~is the least computationally expensive of each of the tracks, and it constrains the model search space enough that ideas are perhaps easier to define and execute. In future iterations of the BabyLM challenge, it could be interesting to provide more specific and constrained \loose~tracks, which focus on particular research directions---for example, LLM-assisted low-resource pretraining, allowing expert annotations during pretraining, or joint text and audio modeling.\looseness=-1

We can also draw insights from the data preprocessing and hyperparameter tuning submissions in particular, and standardize them into the dataset/evaluation pipeline. For example, we could preprocess the data in ways the present challenge has shown to be effective. This could include sorting the data according to the curriculum learning method that yielded performance gains, providing better-starting hyperparameters, and training a baseline with the best architecture. 

Although data quantity was the main focus of this iteration, we may also consider rewarding compute efficiency in the future. 
Many of the most successful submissions consumed a lot of compute by training for many epochs. 
Indeed, the winning submission trained on about as many samples as BERT, despite having a training set only about 3\% as large.
While this finding is interesting, it does little to help achieve our goals in \S\ref{sec:goals}.
Training for hundreds of epochs is not cognitively plausible, and it is does not make it easier and more accessible to test novel training approaches or train models on a university budget.


The evaluation pipeline was built on the existing \texttt{lm-evaluation-harness} repository,\footnote{Originally released at \url{https://github.com/EleutherAI/lm-evaluation-harness}. Note that we based our implementation on the BigScience fork at \url{https://github.com/bigscience-workshop/lm-evaluation-harness}.} but maintaining and updating it for this challenge was no small feat for a single organizer. In future iterations of the challenge, it would be beneficial to have a larger dedicated support team for the evaluations. A dedicated team could also allow us to handle a greater variety of submissions, including those not supported by HuggingFace.


\section{Conclusions} \label{sec:conclusions} 
The BabyLM Challenge encouraged participants to \emph{think small}. We asked: can we improve language modeling on smaller and more cognitively plausible datasets? The submitted systems employed diverse methods, but the most consistent gains came from modified model architectures, new training objectives, principled preprocessing of the pretraining corpora, and hyperparameter searches. In one case, a curriculum learning method resulted in significant improvements. Future work can build on these findings to further improve language modeling for low-resource settings and for cognitive modeling research.

\section*{Acknowledgments}
We would like to thank the participants of the BabyLM Challenge for their valuable contributions---not just their models and papers, but also their contributions to the evaluation pipeline and the reviewing process. 

We would also like to thank the \dynabench team at MLCommons for hosting our leaderboards and integrating our challenge's unique requirements into their implementation. Thanks especially to Max Bartolo, Douwe Kiela, and Hannah Rose Kirk for feedback on earlier iterations of the BabyLM evaluation setup.

\section*{Author Contributions}

\begin{itemize}
    \item \textbf{Original concept:} Alex Warstadt, Leshem Choshen
    \item \textbf{Primary organizers:} Alex Warstadt, Ethan Wilcox, Leshem Choshen, Aaron Mueller, Chengxu Zhuang
    \item \textbf{Pipeline implementation and maintenance:} Aaron Mueller
    \item \textbf{Baseline model training}: Chengxu Zhuang
    \item \textbf{Publicity and communications with participants:} Leshem Choshen, Ethan Wilcox
    \item \textbf{Training dataset compilation:} Alex Warstadt
    \item \textbf{BLiMP Supplement evaluation data creation:} Alex Warstadt
    \item \textbf{\dynabench integration:} Juan Ciro, Rafael Mosquera, Adina Williams
    \item \textbf{Llama 2 evaluation:} Bhargavi Paranjape
    \item \textbf{Guidance on concept and workshop organization:} Ryan Cotterell, Tal Linzen, Adina Williams
    \item \textbf{Reviewing submissions:} Alex Warstadt, Ethan Wilcox, Leshem Choshen, Aaron Mueller, Chengxu Zhuang, Adina Williams
    \item \textbf{Initial draft of findings paper:} Alex Warstadt, Ethan Wilcox, Leshem Choshen, Aaron Mueller, Chengxu Zhuang
    \item \textbf{Editing:} All authors
\end{itemize}

\bibliography{anthology, custom}
\bibliographystyle{acl_natbib}

\appendix
\onecolumn

\section{Data Source Descriptions}\label{app:data_sources}

\paragraph{CHILDES.}
The Child Language Data Exchange System \citep[CHILDES;][]{macwhinney2000childes} is a multilingual database compiling transcriptions from numerous researchers of adult--child interactions in a range of environments, from structured laboratory activities to the home. 
\citet{huebner2021chapter} further process CHILDES, selecting only interactions with American English-speaking children ages 0--6, removing all child utterances, and tokenizing the data. 
The resulting dataset\footnote{\href{https://github.com/phueb/BabyBERTa/blob/master/data/corpora/aochildes.txt}{\url{https://github.com/phueb/BabyBERTa/blob/master/data/corpora/aochildes.txt}}} contains about 5M words.

\paragraph{British National Corpus.}
The BNC \citep{bncconsortium2007british} is a 100M word multi-domain corpus of British English from the second half of the 20$\textsuperscript{th}$ century. 
We select only the dialogue portion of the corpus, totaling about 10M words.

\paragraph{Children's Book Test.}
CBT is a compilation of over a hundred children's books from Project Gutenberg by \citet{hill-2016-cbt}. 
The dataset was originally released with a set of questions for testing named entity prediction, which we do not include in the pretraining data.

\paragraph{Children's Stories Text Corpus.}
This dataset consists of manually selected children's stories from Project Gutenberg. 
It was compiled by \citet{bensaid2021fairytailor} for the development of a story generation system. 

\paragraph{Project Gutenberg.}
The Standardized Project Gutenberg Corpus \citep{gerlach2020standardized} is a curated and preprocessed selection of over 50k literary books in the public domain from Project Gutenberg totaling over 3B tokens.%
\footnote{\href{https://gutenberg.org/}{\url{https://gutenberg.org/}}}
This distribution comes with extensive metadata that allows us to filter texts by language and date.

\paragraph{OpenSubtitles.}
This dataset \citep{lison2016opensubtitles} is a compilation of publicly available subtitles from TV and movies on a third-party website.%
\footnote{\href{http://opensubtitles.org/}{\url{http://opensubtitles.org/}}} 
We use only the English portion.

\paragraph{QED.}
The QCRI Educational Domain Corpus \citep[formerly QCRI AMARA Corpus;][]{abdelali-etal-2014-qed} consists of volunteer-written subtitles for educational videos.
We use only the English portion.

\paragraph{Wikipedia.}
Wikipedia is a volunteer-authored encyclopedia hosted by the Wikimedia Foundation.
We use only the English portion.

\paragraph{Simple English Wikipedia.}
Simple English is classified as a separate language in Wikipedia, thus the texts here are disjoint from those in English Wikipedia.
The texts use shorter sentences and high-frequency vocabulary and avoid idioms.

\paragraph{Switchboard Corpus.}
The Switchboard Corpus \citep{godfrey1992switchboard} is a collection of transcribed telephone conversations between pairs of strangers.
We accessed the text through the Switchboard Dialog Act Corpus \citep{Stolcke-etal:2000}.

\newpage

\begin{minipage}{\textwidth}
\setlength{\parindent}{10mm}

\section{Evaluation Data Details}\label{app:eval-data-size}

As described in \Cref{sec:eval-pipeline}, we filter out evaluation examples that do not have lexical overlap with the \smallstr pretraining corpus. Here, we present the number of training and test examples for each evaluation task after filtering. This allows us to partially control for the confound of the language style of most NLP tasks not aligning well with the pretraining corpus that we constructed. However, we only control for lexical content: other factors, such as sentence length, syntactic complexity, and overall linguistic style, remain distinct between our corpus and these tasks. In the future, it would be helpful for researchers to focus on designing tasks on which both children \emph{and} language models can be reasonably evaluated.

Note, too, that our filtering procedure means that we cannot directly compare results obtained from the BabyLM Challenge to prior evaluations using the full datasets. We use a subset of the training and evaluation examples, and therefore can only compare between models evaluated on our version of these tasks.

\vspace{3em}

    \centering
    \resizebox{0.5\linewidth}{!}{
    \begin{tabular}{clrr}
    \toprule
        & \textbf{Task} & \textbf{$\mid$Train$\mid$} & \textbf{$\mid$Test$\mid$} \\
    \midrule
        \parbox[t]{2mm}{\multirow{12}{*}{\rotatebox[origin=c]{90}{\small{BLiMP}}}} & Anaphor Agreement & -- & 1956 \\
        & Argument Structure & -- & 8248 \\
        & Binding & -- & 6738 \\
        & Control Raising & -- & 4526 \\
        & Determiner-Noun Agreement & -- & 7542 \\
        & Ellipsis & -- & 1732 \\
        & Filler-Gap & -- & 6426 \\
        & Irregular Forms & -- & 1965 \\
        & Island Effects & -- & 2676 \\
        & NPI Licensing & -- & 6586 \\
        & Quantifiers & -- & 3882 \\
        & Subject-Verb Agreement & -- & 5535 \\
    \midrule
        \parbox[t]{2mm}{\multirow{5}{*}{\rotatebox[origin=c]{90}{\small{BLiMP}}}} \parbox[t]{2mm}{\multirow{5}{*}{\rotatebox[origin=c]{90}{\small{Supplement}}}} & Hypernym & -- & 860 \\
        & Question-Answer Congruence (easy) & -- & 64 \\
        & Question-Answer Congruence (tricky) & -- & 165 \\
        & Subject-Auxiliary Inversion & -- & 4099 \\
        & Turn-taking & -- & 280 \\
    \midrule
        \parbox[t]{2mm}{\multirow{11}{*}{\rotatebox[origin=c]{90}{\small{(Super)GLUE}}}} & CoLA & 8164 & 1019 \\
        & SST-2 & 50528 & 508 \\
        & MRPC & 1579 & 177 \\
        & QQP & 243498 & 26889 \\
        & MNLI & 259780 & 6562 \\
        & MNLI-mismatched & 259780 & 6284 \\
        & QNLI & 43917 & 2286 \\
        & RTE & 858 & 99 \\
        & BoolQ & 2072 & 723 \\
        & MultiRC & 4637 & 913 \\
        & WSC & 487 & 83 \\
    \midrule
        \parbox[t]{2mm}{\multirow{11}{*}{\rotatebox[origin=c]{90}{\small{MSGS}}}} & Control Raising (Control) & 6570 & 6731 \\
        & Lexical Content (Control) & 9086 & 9100 \\
        & Main Verb (Control) & 8166 & 8249 \\
        & Relative Position (Control) & 9068 & 9046 \\
        & Syntactic Category (Control) & 8930 & 8824 \\
        & Control Raising--Lexical Content & 6816 & 6910 \\
        & Control Raising--Relative Token Position & 8166 & 8167 \\
        & Main Verb--Lexical Content & 7306 & 7378 \\
        & Main Verb--Relative Token Position & 8177 & 8059 \\
        & Syntactic Category--Lexical Content & 8181 & 7597 \\
        & Syntactic Category--Relative Position & 9159 & 8298 \\
    \bottomrule
    \end{tabular}
    }
    \captionof{table}{Number of training and test examples for each BabyLM evaluation task. We show the number of examples \emph{after} filtering based on the pre-training corpus vocabulary (\Cref{sec:eval-pipeline}).}
    \label{tab:eval-data-size}

\end{minipage}

\section{Examples from the BLiMP Supplement}\label{app:blimp_supplement}
\begin{minipage}{\textwidth}

\begin{tabular}{>{\hangindent=2em}p{0.3\textwidth}>{\hangindent=1em}p{0.3\textwidth} >{\hangindent=1em}p{0.32\textwidth}}
\toprule
    \bf Contrast name & \bf Acceptable sentence & \bf Unacceptable sentence\\
\midrule
    
    \small\contrast{base\_and\_hyponym/ hypernym} & If he is growing herbs, then he is growing plants. & If he is growing herbs, then he is growing basil.\\
    
    \small\contrast{base\_neg\_and\_hypernym\_neg/ converse} & If he isn't growing herbs, that means he isn't growing basil. & If he isn't growing basil, that means he isn't growing herbs.\\
    
    \small\contrast{base\_neg\_and\_hypernym\_neg/ hyponym\_neg} & If he isn't growing herbs, that means he isn't growing basil. & If he isn't growing herbs, that means he isn't growing plants.\\
    
    \small\contrast{hypernym\_and\_base/ converse} & If he is growing basil, that means he is growing herbs. & If he is growing herbs, that means he is growing basil.\\
    
    \small\contrast{hypernym\_and\_base/ other} & If he is growing basil, then he is growing herbs & If he is growing basil, then he is growing flowers.\\
    \small\contrast{hypernym\_and\_other\_neg/ base\_neg} & He is growing basil, therefore he isn't growing flowers. & He is growing basil, therefore he isn't growing herbs.\\
\bottomrule
\end{tabular}
\captionof{table}{Representative examples from the \testsuite{Hypernyms} test suite of the BLiMP supplement.}
\label{tab:hypernym}

\vspace{2.5em}

    \begin{tabular}{ll>{\hangindent=1em}p{0.37\textwidth} >{\hangindent=1em}p{0.37\textwidth}}
    \toprule
        \bf Type & \bf Length & \bf Acceptable dialogue & \bf Unacceptable dialogue\\\midrule
        single & short & David: Should you quit?\textbackslash n Sarah: No, I shouldn't. & David: Should she quit?\textbackslash n Sarah: No, I shouldn't. \\
        single & long & Did they try to finish it on time or not?\textbackslash n No, they didn't. & Did we try to finish it on time or not?\textbackslash n No, they didn't.\\
        double & short & A: Did we say that you finished?\textbackslash n B: Yes, you did. & A: Did you say that you finished?\textbackslash n B: Yes, you did.\\
        double & long & ``Did you say that you will go somewhere after the movie is over?" he asked.\textbackslash n ``No, I didn't," she said.  & ``Did you say that you will go somewhere after the movie is over?" he asked.\textbackslash n ``No, you didn't," she said.\\
    \bottomrule
    \end{tabular}
    \captionof{table}{Representative examples from the \testsuite{Turn-Taking} test suite of the BLiMP supplement.}
    \label{tab:turntaking1}

\vspace{2.5em}

    \resizebox{\textwidth}{!}{%
    \begin{tabular}{ll>{\hangindent=1em}p{0.39\textwidth} >{\hangindent=1em}p{0.39\textwidth}}
    \toprule
        \bf Contrast name & \bf Dif. & \bf Acceptable dialogue & \bf Unacceptable dialogue\\\midrule
        \small\contrast{animate vs. inanimate} & easy & A: What did you purchase?\n~B: Bread. & A: What did you purchase?\n~B: David.\\
        \small\contrast{inanimate vs. animate} & easy & ``Who played the piano?" he asked. ``A teacher played the piano," she said. & ``Who played the piano?" he asked. ``A car played the piano," she said.\\
        \small\contrast{loc vs. NP} & easy & David: Where did you put it?\n~Sarah: Behind the sofa. & David: Where did you put it?\n~ Sarah: Eggs. \\
        \small\contrast{animate vs. inanimate} & tricky & David: Who mopped?\n~Sarah: A doctor. & David: Who mopped?\n~Sarah: The tiles. \\
        \small\contrast{loc vs. NP} & tricky & A: Where were you reading?\n~B: By the lake. & A: Where were you reading?\n~B: An essay.\\
        \small\contrast{temp vs. NP} & tricky & When did you eat?\n~Several minutes ago. & When did you eat?\n~Dinner. \\
        \small\contrast{expl vs. NP} & tricky & ``Why were you reading?" he asked. ``For fun," she said. & ``Why were you reading?" he asked. ``A book," she said. \\
        \small\contrast{num vs. NP} & tricky & A: How many do you teach?\n~B: A few. & A: How many do you teach?\n~B: History. \\
        \small\contrast{manner vs. NP} & tricky & David: How did you vacuum?\n~Sarah: I vacuumed quickly. & David: How did you vacuum?\n~Sarah: I vacuumed the patio. \\
        
    \bottomrule
    \end{tabular}
    }
    \captionof{table}{Representative examples from the \testsuite{Question-Answer Congruence} test suite of the BLiMP supplement.}
    \label{tab:turntaking2}

\end{minipage}

\section{Subtask Results}
Here, we present a more detailed breakdown of results by subtask. Each task has a subsection containing a table where results are described, as well as a textual description containing and overview of the main takeaways for each task.

\subsection{MSGS}
\setlength{\parindent}{5mm}
Matthews correlation coefficients on MSGS (Table~\ref{tab:msgs_subtasks}) were largely negative, indicating that language models trained at this scale tend to prefer surface features over linguistic features in ambiguous contexts. However, certain models demonstrated a much stronger preference for linguistic features in specific contexts: ELC-BERT showed high positive scores on average (sometimes significantly higher than Llama 2), as did Contextualizer. This shows us that architectural modifications can significantly improve scores, as can princpled approaches to curriculum learning.

In general, comparable models trained on the \strict corpus have higher MCCs than those trained on the \smallstr corpus, but not always. This suggests that, while more pretraining data generally lead to stronger syntactic inductive biases, these preferences may depend on the features being compared, and that this will not always be the case depending on the architecture used.
\vspace{1em}

\def\msgstask#1{\rotatebox[origin=c]{90}{\parbox[t]{6em}{\small\textbf{#1}}}}
{\centering
\resizebox{\textwidth}{!}{%
\begin{tabular}{lllllllll}
    \toprule
        \textbf{} & \textbf{Model} & 
        \msgstask{Macro\\average} & 
        \msgstask{Ctl--Raising /\\Lexical content} & 
        \msgstask{Ctl--Raising /\\Relative position} & 
        \msgstask{Main verb /\\Lexical content} & 
        \msgstask{Main verb /\\Relative position} & 
        \msgstask{Syntactic cat. /\\Lexical content} & 
        \msgstask{Syntactic cat. /\\Relative position} \\ 
    \midrule
        ~ & Llama 2 \citep{Llama2} & \best{-0.24} & \bestall{\-0.93} & \best{\-0.23} & \best{-0.77} & -0.96 & -0.19 & -0.74 \\ 
        ~ & RoBERTa-base \citep{liu-etal-2019-roberta} & -0.37 & \-0.46 & -0.58 & -0.95 & \best{-0.94} & \best{\-0.36} & \best{-0.57} \\ 
    \midrule
        \parbox[t]{2mm}{\multirow{4}{*}{\rotatebox[origin=c]{90}{\small{Strict}}}} 
        & ELC-BERT \citep{babylm2023notall} & \best{\-0.10} & -0.51 & \best{-0.46} & \best{\-0.71} & \bestall{\-0.97} & \bestall{\-0.46} & -0.53 \\ 
        ~ & Boot-BERT \citep{babylm2023meanberts} & -0.22 & \-0.37 & -0.77 & -0.99 & \-0.96 & -0.34 & -0.58 \\ 
        ~ & McGill \citep{babylm2023mcgill} & -0.35 & \best{\-0.65} & -0.70 & -0.99 & -0.73 & \-0.17 & \best{-0.49} \\ 
        ~ & \emph{Best Baseline (OPT-125M)} & -0.39 & \-0.35 & -0.70 & -0.76 & -0.99 & \-0.34 & -0.60 \\ 
    \midrule
        \parbox[t]{2mm}{\multirow{4}{*}{\rotatebox[origin=c]{90}{\small{Strict-small}}}} 
        & ELC-BERT \citep{babylm2023notall} & \best{-0.01} & \-0.02 & -0.71 & \bestall{\-0.95} & \best{\-0.50} & -0.26 & -0.59 \\ 
        ~ & MLSM \citep{babylm2023cogmemlm} & -0.37 & \best{\-0.31} & -0.56 & -0.99 & -0.49 & -0.03 & -0.44 \\ 
        ~ & McGill \citep{babylm2023mcgill} & -0.60 & -0.68 & \best{-0.37} & -1.00 & -0.79 & -0.35 & \best{-0.42} \\ 
        ~ & \emph{Best Baseline (OPT-125M)} & -0.45 & \-0.00 & -0.70 & -0.72 & -0.77 & \best{\-0.13} & -0.68 \\ 
    \midrule
        \parbox[t]{2mm}{\multirow{3}{*}{\rotatebox[origin=c]{90}{\small{Loose}}}} 
        & Contextualizer \citep{babylm2023contextualizer} & \bestall{\-0.24} & \bestall{\-0.88} & \bestall{\-0.71} & \best{-0.32} & \best{\-0.30} & \best{\-0.21} & \bestall{-0.35} \\ 
        ~ & McGill \citep{babylm2023mcgill} & -0.75 & -0.56 & -0.97 & -0.99 & -0.86 & -0.66 & -0.46 \\ 
        ~ & BabyStories \citep{babylm2023babystories} & -0.71 & -0.24 & -0.99 & -0.99 & -0.99 & -0.23 & -0.78 \\ 
    \bottomrule
\end{tabular}}}
\captionof{table}{MSGS results for each ambiguous subtask for the top performing models (by overall score) from each track, as well as baselines and skylines. MCC (i.e., linguistic bias score) results presented, truncated to two decimal places.}
\label{tab:msgs_subtasks}

\subsection{BLiMP}
\setlength{\parindent}{5mm}
Accuracies on BLiMP (Table~\ref{tab:blimp_subtasks}) show that bigger models do not, as a rule, perform better on targeted grammatical evaluation. 
RoBERTa is the best-performing skyline model, despite that Llama 2 has orders-of-magnitude more parameters and was trained on significantly more data. 
Among the BabyLM submissions, Boot-BERT generally performs best, with ELC-BERT and McGill's submission also performing well in general on the \strict and \smallstr tracks. 
ELC-BERT and Boot-BERT are both based on LTG-BERT \citep{samuel2023trained}, suggesting that this architecture is a good starting point for pretraining on developmentally plausible amounts of linguistic input. 

Analyzing specific test suites, we see that unsurprisingly that models in all tracks typically perform best on agreement phenomena, though we find surprisingly high variability on \testsuite{anaphor agreement}.
\citet{warstadt-etal-2020-blimp-benchmark--} reported that \testsuite{island effects} and \testsuite{quantifiers} were the two most difficult test cases.
We find that the best BabyLM submissions actually outperform Llama by a wide margin on \testsuite{island effects}. 
However, \testsuite{quantifiers}, on which most models achieve very consistent and mediocre results, is the one test suite on which the Llama 2 skyline is stronger.

\subsection{BLiMP Supplement}
\setlength{\parindent}{5mm}
Accuracies on the BLiMP supplement tasks (Table~\ref{tab:supplement_subtasks}) demonstrate similar trends as those in the BLiMP tasks.
As these individual test suites are new to this task, these fine-grained results are of particular interest.
We find that the \testsuite{hypernym} test suite is clearly beyond the ability of language models.
All models including the skylines perform very close to chance, suggesting either that their preferences are virtually random guessing, or they show systematic biases that essentially cancel out due to counterbalancing in the test data.
However, we hesitate to conclude that these models have no knowledge of lexical entailment relations for two reasons:
First, these test sentences are somewhat unnatural logical statements which are out-of-domain for the models, and
second, there is less reason \emph{a priori} to think that logically invalid statements have lower probability than valid statements.

Among the \testsuite{question--answer congruence} test suites, we do indeed find that the ``tricky'' examples are far more difficult than the ``easy'' ones.
The ``tricky'' set is highly discriminative, due probably to its adversarial nature, telling us that most models are easily fooled by locally coherent distractor answers and pay too little attention to cross-sentential long-distance dependency between a \emph{wh}-word and a congruent answer.
Only the top-performing models in the \strict track score better than chance, and the RoBERTa skyline outperforms all models by a wide margin.

The tests for \testsuite{subject--auxiliary inversion} are relatively easy, with the best models reaching near-perfect accuracy.
\testsuite{Turn taking} is highly discriminative, with some models performing at or near chance, while the best model achieves accuracy over 90\%.
Again, ELC-BERT outperforms the skylines.
This may be due in part to the fact that transcribed dialogue is a relatively large proportion of the BabyLM training data, compared to the training data for typical pretrained language models.

\def\supptask#1{\rotatebox[origin=c]{90}{\parbox[t]{6em}{\small\textbf{#1}}}}
{\centering
\begin{tabular}{llllllll}
\toprule
    \textbf{} & \textbf{Model} & 
    \supptask{Macro\\average} & 
    \supptask{Hypernym} & 
    \supptask{Q--A congruence\\(easy)} & 
    \supptask{Q--A congruence\\(tricky)} & 
    \supptask{Subject--aux\\inversion} & 
    \supptask{Turn taking} \\ 
\midrule
    ~ & Llama 2 & 0.74 & \bestall{0.50} & 0.85 & 0.63 & 0.91 & \best{0.83} \\ 
    ~ & RoBERTa & \best{0.75} & 0.48 & \bestall{0.87} & \bestall{0.72} & \bestall{0.98} & 0.73 \\
\midrule
    \parbox[t]{2mm}{\multirow{4}{*}{\rotatebox[origin=c]{90}{\small{Strict}}}} 
    & ELC-BERT & \bestall{0.76} & \best{0.47} & \bestall{0.85} & \bestall{0.63} & 0.94 & \bestall{0.92} \\ 
    ~ & Boot-BERT & 0.72 & 0.45 & 0.75 & 0.58 & \bestall{0.96} & 0.86 \\ 
    ~ & McGill & 0.71 & 0.46 & 0.84 & 0.58 & 0.82 & 0.83 \\ 
    ~ & OPT & 0.67 & 0.46 & 0.76 & 0.47 & 0.85 & 0.82 \\ 
\midrule
    \parbox[t]{2mm}{\multirow{4}{*}{\rotatebox[origin=c]{90}{\small{Strict-small}}}} 
    & ELC-BERT & \best{0.67} & 0.48 & 0.68 & \best{0.44} & \best{0.88} & \best{0.83} \\ 
    ~ & MLSM & 0.57 & 0.47 & 0.70 & 0.33 & 0.82 & 0.52 \\ 
    ~ & McGill & 0.58 & 0.49 & \best{0.73} & 0.35 & 0.77 & 0.57 \\ 
    ~ & OPT & 0.52 & \bestall{0.50} & 0.54 & 0.31 & 0.70 & 0.57 \\ 
\midrule
    \parbox[t]{2mm}{\multirow{3}{*}{\rotatebox[origin=c]{90}{\small{Loose}}}} 
    & Contextualizer & \best{0.63} & 0.47 & \best{0.73} & 0.42 & \best{0.91} & 0.62 \\ 
    ~ & McGill & 0.56 & \best{0.49} & 0.64 & 0.29 & 0.80 & 0.61 \\ 
    ~ & BabyStories & 0.64 & \best{0.49} & 0.71 & \best{0.50} & 0.79 & \best{0.73} \\ 
\bottomrule
\end{tabular}
\captionof{table}{BLiMP Supplement accuracies for each subtask for the top performing systems (by overall score), best baseline, and skylines. For each subtask, we mark the \best{best} performing system for each track, and the \bestall{best} non-skyline and \bestall{best} performing system overall.}
\label{tab:supplement_subtasks}
}

\subsection{GLUE/SuperGLUE}
Scores on (Super)GLUE tasks (Table~\ref{tab:glue_subtasks}) show that ELC-BERT is generally the best-performing system in both the \strict and \smallstr tracks, and that Boot-BERT is also highly effective in the \strict track. Contextualizer also performs well. This largely confirms findings from the BLiMP and BLiMP Supplement tasks: LTG-BERT is an effective architecture for pretraining on smaller corpora, and curriculum learning can improve performance over a naïve corpus ordering.

\def\gluetask#1{\rotatebox[origin=c]{90}{\parbox[t]{3em}{\small\textbf{#1}}}}
\resizebox{\textwidth}{!}{%
\begin{tabular}{llllllllllllll}
\toprule
    ~ & \textbf{Model} & 
    \gluetask{Macro\\average} & 
    \gluetask{CoLA} & 
    \gluetask{SST-2} & 
    \gluetask{MRPC} & 
    \gluetask{QQP} & 
    \gluetask{MNLI} & 
    \gluetask{MNLI-mm} & 
    \gluetask{QNLI} & 
    \gluetask{RTE} & 
    \gluetask{BoolQ} & 
    \gluetask{MultiRC} & 
    \gluetask{WSC} \\ 
\midrule
    ~ & Llama 2 & \bestall{0.83} & \bestall{0.63} & \bestall{0.95} & 0.87 & 0.81 & 0.85 & \bestall{0.87} & 0.89 & \bestall{0.81} & \bestall{0.85} & \bestall{0.86} & \bestall{0.75} \\ 
    ~ & RoBERTa & 0.78 & 0.62 & 0.93 & \best{0.88} & \best{0.87} & \bestall{0.86} & 0.85 & \bestall{0.92} & 0.61 & 0.76 & 0.68 & 0.61 \\ 
\midrule
    \parbox[t]{2mm}{\multirow{4}{*}{\rotatebox[origin=c]{90}{\small{Strict}}}} & ELC-BERT & \bestall{0.78} & \bestall{0.59} & \bestall{0.92} & \bestall{0.90} & \bestall{0.88} & 0.84 & 0.83 & 0.89 & 0.64 & \bestall{0.73} & 0.72 & \bestall{0.62} \\ 
    ~ & Boot-BERT & \bestall{0.78} & 0.57 & \bestall{0.92} & 0.89 & \bestall{0.88} & \bestall{0.85} & \bestall{0.84} & \bestall{0.91} & \bestall{0.65} & 0.72 & \bestall{0.73} & 0.61 \\ 
    ~ & McGill & 0.72 & 0.49 & 0.89 & 0.83 & 0.86 & 0.79 & 0.79 & 0.84 & 0.53 & 0.66 & 0.65 & 0.61 \\ 
    ~ & OPT & 0.70 & 0.36 & 0.88 & 0.82 & 0.83 & 0.76 & 0.77 & 0.83 & 0.63 & 0.66 & 0.60 & 0.54 \\ 
\midrule
    \parbox[t]{2mm}{\multirow{4}{*}{\rotatebox[origin=c]{90}{\small{Strict-small}}}} & ELC-BERT & \best{0.73} & \best{0.47} & 0.86 & \best{0.87} & \best{0.86} & \best{0.78} & \best{0.79} & \best{0.84} & \best{0.60} & \best{0.69} & \best{0.68} & \bestall{0.62} \\ 
    ~ & MLSM & 0.70 & 0.41 & \best{0.90} & 0.78 & 0.85 & 0.75 & 0.76 & 0.82 & 0.59 & 0.66 & 0.58 & 0.61 \\ 
    ~ & McGill & 0.69 & 0.41 & 0.87 & 0.79 & 0.81 & 0.73 & 0.74 & 0.79 & 0.54 & 0.66 & 0.62 & 0.61 \\ 
    ~ & OPT & 0.62 & 0.15 & 0.84 & 0.74 & 0.78 & 0.67 & 0.69 & 0.65 & 0.55 & 0.65 & 0.51 & 0.59 \\ 
\midrule
    \parbox[t]{2mm}{\multirow{3}{*}{\rotatebox[origin=c]{90}{\small{Loose}}}} & Contextualizer & \best{0.72} & \best{0.56} & \best{0.90} & \best{0.83} & \best{0.85} & \best{0.77} & \best{0.78} & \best{0.83} & \best{0.53} & \best{0.68} & \best{0.64} & 0.59 \\ 
    ~ & McGill & 0.68 & 0.37 & 0.88 & 0.77 & 0.83 & 0.73 & 0.75 & 0.78 & 0.49 & 0.67 & 0.60 & \best{0.61} \\ 
    ~ & BabyStories & 0.60 & 0.00 & 0.84 & 0.82 & 0.66 & 0.59 & 0.64 & 0.79 & \best{0.53} & 0.67 & 0.46 & \best{0.61} \\ 
\bottomrule
\end{tabular}}
\captionof{table}{(Super)GLUE results for each subtask for the top performing systems (by overall score), best baseline, and skylines. For each subtask, we mark the \best{best} performing system for each track, and the \bestall{best} non-skyline and \bestall{best} performing system overall.}
\label{tab:glue_subtasks}

\section{Age of Acquisition Prediction Results}\label{app:aoa_eval}
\setlength{\parindent}{5mm}
Here, we present scores, separated by track, for each model that evaluated on the age of acquisition (AoA) prediction task (Table~\ref{tab:aoa_eval}). We also compare to the best-performing baseline within each track, as in \Cref{tab:result}.

Almost all submissions which evaluated on the AoA prediction task were in the \smallstr track. Here, no model achieved closer predictions than the OPT-125M baseline, though many got very close. In the \strict track, BabyStories achieved very close scores to the OPT-125M baseline.

\vspace{1em}

{\centering
\resizebox{\textwidth}{!}{%
    \begin{tabular}{clrrrr}
    \toprule
        & \multirow{2}{*}{\textbf{Model}} & \multicolumn{4}{c}{\bf Mean average deviation $\downarrow$}\\
        
        & & \bf Overall & \bf Nouns &\bf Predicates & \bf  Function Words \\
    \midrule
        \parbox[t]{2mm}{\multirow{2}{*}{\rotatebox[origin=c]{90}{\small{Strict}}}} & BabyStories (GPT2-Large-PPO)~\cite{babylm2023babystories} & 2.05 & 1.98 & \underline{\textbf{1.82}} & 2.63 \\
         & \emph{Best Baseline (OPT-125M)} & \underline{\textbf{2.04}} & \underline{\textbf{1.97}} & 1.83 & \underline{\textbf{2.61}} \\
    \midrule
        \parbox[t]{2mm}{\multirow{8}{*}{\rotatebox[origin=c]{90}{\small{Strict-Small}}}}  & GPT-Wee (16k (cu.))~\citep{babylm2023gptwee} & 2.06 & 2.00 & 1.83 & 2.58 \\
         & Bebeshka \citep{babylm2023miniminds} & 2.06 & 1.98 & 1.84 & 2.66 \\
         & Zlata \citep{babylm2023miniminds} & 2.07 & 1.99 & 1.83 & 2.67 \\
         & Too Much Information \citep{babylm2023toomuch} & 2.05 & 1.99 & 1.85 & 2.58 \\
         & Mmi01 (\textsc{Rarity})~\cite{babylm2023mmi01} & 2.05 & \underline{\textbf{1.97}} & 1.85 & 2.64 \\
         & Baby Llama~\citep{babylm2023babyllama} & 2.06 & 1.99 & 1.84 & 2.63 \\
         & Lil-Bevo-X~\citep{babylm2023lilbevo} & 2.05 & 1.99 & 1.85 & 2.59 \\
         & \emph{Best Baseline (OPT-125M)} & \underline{\textbf{2.03}} & 1.98 & \underline{\textbf{1.81}} & \underline{\textbf{2.57}} \\
    \bottomrule
    \end{tabular}}}
    \captionof{table}{Mean average deviation (MAD) in months across cross-validation folds when predicting the age of acquisition of words. Lower MAD scores are better. We present all systems that evaluated on AoA prediction, as well as the baseline model with the best scores per track. We \underline{\textbf{bold}} the highest-scoring system for each task within each track.}
    \label{tab:aoa_eval}
    \vspace{-10pt}


\vspace{2.5em}

\section{Summary of Each Submission}
\label{app:summaries}

\paragraph{GPT-wee \citep{babylm2023gptwee}.} This paper tests various approaches to reordering the examples based on word and sentence statistics. The motivation comes from usage-based linguistics and the idea that frequent lexical items, such as phrases or common groups of words, are learned early (rather than words, for instance). They also find that training more---up to 10 epochs---helps, and that a medium-sized model might be as good as larger models.

\paragraph{Tiny Language Models with Multiplex Networks \citep{babylm2023tinylang}.} This approach leverages multimodal data (including text/visual data and sensorimotor data) as part of the embeddings to an ELECTRA language model. The proposed models are very small (as few as 7M parameters) and perform well on BLiMP. For reference, the baseline models contain 125M to 220M parameters.

\paragraph{Mini Minds \citep{babylm2023miniminds}.} This submission explores how scaling down models (in terms of number of parameters) can help in low-data settings. The authors conduct a parameter search for scaled-down versions of GPT-2 and RoBERTa, and find that optimal models have around a 2-to-1 ratio of attention heads to layers. They train two models and find that they perform about as well as larger parameter count models on GLUE. Furthermore, the authors test their models on an ethical reasoning benchmark and find that the small models perform about as well as models which have about ten times the parameters.

\paragraph{Grammar induction pretraining \citep{babylm2023grammar}.}
This submission introduces syntactic bias into the static token embeddings of an LM. An unsupervised grammar induction system is trained on a 1-million word subset of the \smallstr corpus, and the resulting static token embeddings are used to initialize the LM token embeddings. Although the results improve over the BabyLM \smallstr baseline, similar improvements are observed with a custom baseline model using randomly initialized token embeddings. Thus, there is no evidence that the grammar induction step had a positive impact on LM results.

\paragraph{ChapGTP \citep{babylm2023chapgtp}.} This work explores how targeted data augmentation can improve the performance of masked language models in the \smallstr track. The authors used regex patterns to extract common phrases from the GLUE tasks and then used these patterns to generate follow-up questions that served as additional training data. They also found that increasing the training epochs up to 200 epochs continues to help performance.

\paragraph{BabyBerta+ \citep{babylm2023pennbgu}.} The submission replicates the BabyBERTa training setup \citep{huebner2021babyberta} and tests its ability after pretraining on the \smallstr corpus. They find that a small model trained on many epochs keeps improving and becomes better than baseline models in grammatical aspects, but not downstream tasks. 

\paragraph{Keeping Training Simple for BabyLMs \citep{babylm2023toomuch}.} This paper proposes a variety of complexity metrics for reordering the BabyLM \smallstr data from simple to complex. Compared to no curricula and reversed curricula, the proposed curricula do not result in consistent performance improvements on the BabyLM evaluation tasks. However, reducing the context length to 32 (from the baselines' 128) results in significant and consistent performance improvements.

\paragraph{Can Training Neural Language Models on a Curriculum with Developmentally Plausible Data Improve Alignment with Human Reading Behavior? \cite{babylm2023cantrain}.} This paper explores surprisal-based curricula for pretraining on the \smallstr dataset of the BabyLM challenge. The authors use an ensemble of LSTM ``teacher'' models to rank sentences by average surprisal, on which a final OPT model is trained. Results are mixed. The authors find that their model does not outperform a random baseline. However, when this model is further trained on the randomly-ordered training dataset after training on the curriculum-ordered data, it does beat the baseline. As an additional analysis, the authors investigate the ability of their model to predict human reading times for syntactically complex sentences, finding that the model is not particularly good at the task, but that it is about equivalent to baselines which are trained on much larger datasets.

\paragraph{CLIMB \citep{babylm2023climb}.}
This submission presents a thorough comparison of different approaches to curriculum learning in the \smallstr setting. They consider three main criteria for curriculum design: the size of the input vocabulary, the difficulty of the training sample, and the size of the output space for MLM prediction. They conduct experiments exploring eight different curricula sorted into these three main approaches. While there are many small differences in performance among these settings, curricula provide no consistent improvements over more naive training algorithms.

\paragraph{Acquiring Linguistic Knowledge from Multimodal Input \citep{babylm2023acqling}.} The authors explored whether vision-language co-training helps the learning of linguistic knowledge. They trained models on Wiki texts with images using the state-of-the-art multi-modality model (FLAVA). After varying the amount of training data and how many images are used, the authors found that visual input only provides a slight improvement on grammar benchmarks for 10M-word training, but not for 100M-word training.

\paragraph{GPT-like Models are Bad Babies \citep{babylm2023gptlike}.} This paper trains a decoder-only model, trying different hyperparameters, including reordering the training data by different orders (based on cues which did not improve over regular shuffling), different sizes, layer widths, among other features. The main focus of the paper is to test if models that perform better on BabyLM evaluation tasks are also better at modeling reading difficulty in humans. Surprisingly, models performing better on BabyLM tasks performed \textit{less} well in modeling reading difficulty.

\paragraph{Baby's CoThought \citep{babylm2023babycot}.} This system leverages a large language model, GPT-3.5-Turbo, to reformat semantically unrelated sentences into cohesive paragraphs. In low-data settings, this approach can form better training examples for language models; the proposed approach results in improvements across BLiMP tasks, though performance is not significantly different on (Super)GLUE or MSGS. Note that the LLM is trained on far more than 100M words, so this submission technically does not qualify under any track. However, this method does improve the sample efficiency of the student model, and it aids our understanding of what types of data are best for supervising smaller language models.

\paragraph{ToddlerBERTa \citep{babylm2023todberta}.} This paper conducts a thorough hyperparameter investigation of the BabyBERTa model, exploring different options for model sizes and training algorithms. The author finds that larger models tend to perform better.

\paragraph{CogMemLM \cite{babylm2023cogmemlm}.} 
This work explores an approach to word segmentation and tokenization that is intended to model vocabulary growth during learning. 
A vocabulary is cumulatively built using a cognitively-inspired model of word segmentation, in which strings are split into chunks based on an activation weight which changes throughout training depending on how often the chunk is observed together. 
While the approach achieves consistent improvements over the BabyLM \strict baseline results, it is not clear whether these improvements are due to the segmentation scheme or other hyperparameter modifications.

\paragraph{BabyStories \citep{babylm2023babystories}.} This paper investigates how reinforcement learning from human feedback (RLHF) improves the performance of causal language models pretrained on small scales of datasets. The authors report that models finetuned by RLHF on short stories yield better performance on language understanding benchmarks, though this improvement is only observed on larger models. Their findings suggest that benefiting from RLHF requires a large number of trainable parameters.

\paragraph{Byte-ranked Curriculum Learning \citep{babylm2023byterank}.} This paper proposes a curriculum learning approach for reordering data based on non-linguistic metrics. Specifically, they choose the order in which datasets are shown to the model starting from the minimal amount of bytes per sentence and going up. This happens to also start from spoken data and follow with text data later. The paper also shows that a larger model as well as more epochs improves the results.

\paragraph{McGill BabyLM Submission \citep{babylm2023mcgill}.} This paper finds that changes to the data format have large positive impacts. Specifically, not using sequence packing, using sentences and not documents as examples, not truncating, and reducing maximum sequence length are each highly effective. By contrast, adding supervision from POS tags and using unsupervised syntactic induction have negligible impact.

\paragraph{Mean BERTS make erratic language teachers \citep{babylm2023meanberts}.} This submission presents Boot-BERT, a latent bootstrapping approach to language modeling in low resource settings. In the latent bootstrapping set-up, a student model is trained to produce predictions over words as well as to match contextualized embeddings from a teacher model. In turn, the teacher’s embeddings are obtained via a moving average of the student’s. The authors use LTG-BERT \citep{samuel2023trained} as an encoder backbone, as well as for a baseline.\footnote{As described in \S\ref{sec:winning-submissions}, LTG-BERT makes multiple modifications to the standard Transformer encoder architecture: additional layer normalization \citep{shleifer2022normformer}, GEGLU feed-forward modules \citep{shazeer2020glu}, disentangled attention following DeBERTa \citep{he2021deberta}, and scaled weight initialization following \cite{nguyen-salazar-2019-transformers}.} They find that their Boot-BERT outperforms LTG-BERT for some of the BabyLM tasks, including GLUE for both the \strict and \smallstr tracks.

\paragraph{Every Layer Counts BERT (ELC-BERT) \citep{babylm2023notall}.}
This submission takes as its starting point the very effective LTG-BERT architecture from \citet{samuel2023trained} and modifies it such that the input to each layer is a weighted sum of the outputs of all previous layers, where the weights can be learned but also biased by initialization. Several variations are explored, including equal initial weights, and initial weights biased towards the previous layer. Results on BabyLM evaluations do not strongly suggest that any one variant is clearly better than the LTG-BERT baseline, though all models perform significantly better than the BabyLM RoBERTa baseline. Additionally, inspection of the learned weights for combining previous layer outputs suggests that the most important outputs are from the previous few layers and the static embedding layer.

\paragraph{WhisBERT \citep{babylm2023whisbert}.} In this submission, the authors explore whether text-and-audio co-training helps model performance on BLiMP tasks. After pretraining a multi-modal model (FLAVA) on 100M words with or without their corresponding word-aligned speech, they find that the speech-augmented model outperforms the text-only model on 11 out of 17 grammatical tasks. 

\paragraph{Surprisal-based active curriculum learning \citep{babylm2023surporc}.} This submission combines curriculum and active learning to schedule training order for models. The authors use n-gram surprisals to determine the sentences with the highest surprisal and then train their models on structurally similar examples to these high-surprisal sentences. Models with active curriculum learning show noticeable performance gains in (Super)GLUE but underperform the models without such learning on MSGS.

\paragraph{Linguistically Motivated Curriculum Learning \citep{babylm2023mmi01}.} This submission tests 6 linguistic metrics of complexity as curriculum learning approaches. On the \smallstr track, this approach succeeds in finding improvements over training on the whole corpus in a random order.

\paragraph{Baby Llama \citep{babylm2023babyllama}.} This submission proposes a knowledge distillation approach with two teacher models (a 300M-parameter Llama model and 700M-parameter GPT-2 model) trained on the \smallstr corpus. These are distilled into a 58M-parameter Llama model called Baby Llama. The proposed model outperforms the BabyLM baselines, the teacher LMs, and a 58M-parameter Llama model trained from scratch on the \smallstr data without distillation.

\paragraph{Curriculum learning based on sentence complexity approximating language acquisition \citep{babylm2023conllst}.} This submission assesses the impact of curriculum learning based on sentence complexity within the context of the \smallstr task. The authors order training data based on three sentence-level complexity metrics: number of tokens, number of constituents, and max depth of the sentences’ dependency parse. They find that the dependency-based ranking leads to better models, however, all curriculum-based models underperform a random baseline.

\paragraph{Masked Latent Semantic Modeling \citep{babylm2023bettertogether}.}
This paper adopts a method from \citet{berend2023masked} called Masked Latent Semantic Modeling (MLSM) in which the target output distribution can be transformed from a one-hot distribution over the vocabulary into a sparse distribution over latent ``semantic property'' vectors. Then, the same kind of student-teacher optimization as in knowledge distillation is applied using this modified output distribution instead of the full vocabulary. MLSM on its own is found to lead to degradation in BLiMP performance, although combining MLSM with typical MLM training in a multitask setting leads to similar performance as MLM training alone.

\paragraph{Lil-Bevo \citep{babylm2023lilbevo}.} This paper offered submissions to both \smallstr and \strict tracks and used three design choices for LM training: (i) initially pretraining on music data, following work on transfer learning \citep{papadimitriou-jurafsky-2020-learning2}, which suggested that musical structure may form a reasonable basis upon which to learn language structure; (ii) subsequently using a training curriculum starting from shorter sequences (128) before moving to longer ones (512), following insights from \citet{press-etal-2021-shortformer}, and (iii) masking critical tokens necessary to perform some of the BLiMP subtasks (e.g., masking ``not'' for NPI-licensing). Taking final results into consideration alongside ablations, this team found that sequence length matters, music pretraining may help a little, and targeted MLM training seems to help (but only for some BLiMP subtasks, including NPI licensing and Argument Structure).

\paragraph{Contextualizer \citep{babylm2023contextualizer}.} This paper sorts the corpora in the training dataset loosely based on their age of acquisition and reading difficulty. The authors then introduce techniques to begin and end the training with padding-separated datasets sorted from easy to hard, while the middle of the training employs a noisier padding and sorting strategy to improve the model's robustness. The final model performs similarly to its counterpart pretrained with thousands of times more data.

\paragraph{Implicit Structure Building \citep{babylm2023implicitstructure}.} This submission introduces an unsupervised hierarchical bias into the transformer. The approach shows that such structural bias with StructFormer improves over the classic MLM Transformer approach. Improvements are not consistent across scenarios: the model excels in single-sentence or syntactic evaluation tasks, but less so in semantic tasks with multi-sentence inputs.

\paragraph{Pretraining LLMs using human-like development data \citep{babylm2023llmsdev}.} This submission trains RoBERTa, DistilBERT, and GPT-2 models on the \strict and \smallstr data. They find that training DistilBERT for 60 epochs is better than 20 epochs. They also claim that the performance of the baseline RoBERTa model may not be replicable across random initializations and that hyperparameter searches should be more thorough to hedge against such outlier models.

\paragraph{On the Effect of Curriculum Learning with Developmental Data for Grammar Acquisition \citep{babylm2023curriculum}.} This submission explores the effect of curriculum learning, using BabyBERTa models, on the \smallstr data track. The authors contrast three types of curriculum learning: one that orders input by word frequency; one by sequence entropy; and one by increasing context length. They find that neither of these methods produces results above a baseline random presentation. In a series of follow-up experiments, the authors verify that model performance is linked to the amount of exposure to transcribed speech data and suggest that speech data is a good foundation for curriculum learning.

\paragraph{Difficulty-based Sentence Reordering \citep{babylm2023gpt2opt}.}
This study explores two broad approaches to dataset preprocessing to improve LM training in the 10M-word setting: data reordering (curriculum learning) and data cleaning. Results show that reordering a subset of the data by sentence difficulty may lead to marginal improvements, as long the local coherence of the samples is not damaged too greatly. However, the clearest improvements come from cleaning the data of incoherent, ungrammatical, or non-linguistic strings.

\newpage

    
\section{Results Broken Down by GLUE / BLiMP Subtask}


\vspace{2.5em}

\includegraphics[width=\linewidth]{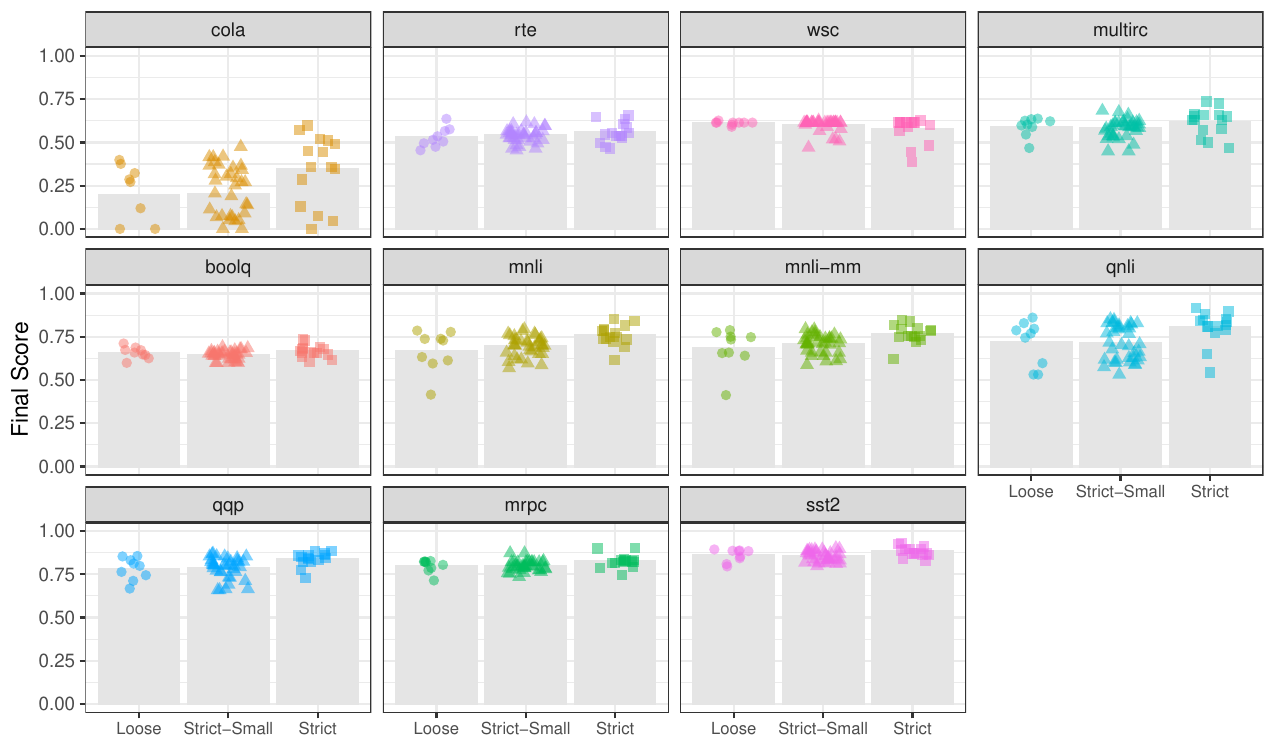}
    \captionof{figure}{\textbf{Submission Results by GLUE subtask:} Points show the performance of each submission. Gray bars show the across-submission average in each category.}
    \label{fig:results-glue}


\vspace{2.5em}
    
    \includegraphics[width=\linewidth]{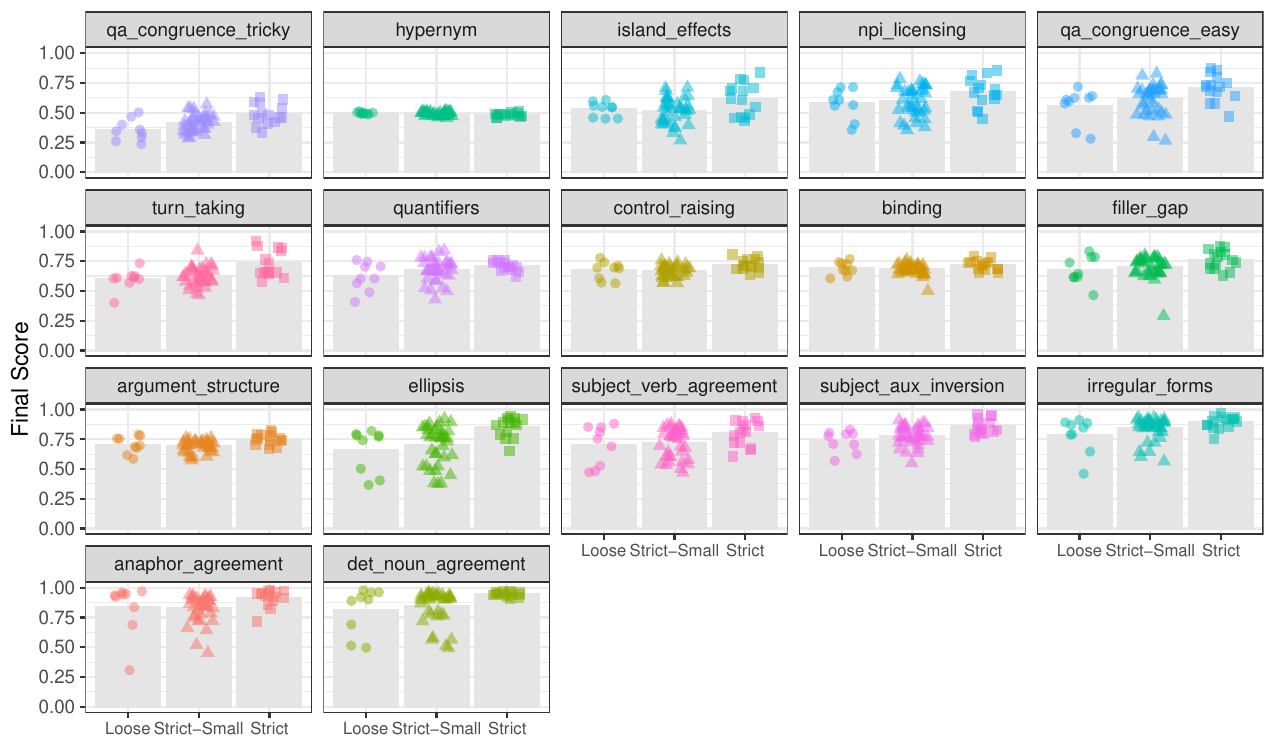}
    \captionof{figure}{\textbf{Submission Results by BLiMP subtask:} Points show the performance of each submission. Gray bars show the across-submission average in each category. }
    \label{fig:results-blimp}


\end{document}